\documentclass[11pt]{article}
\usepackage{acl2015}
\usepackage{times}
\usepackage[np]{numprint}
\usepackage{url}
\usepackage{latexsym}
\usepackage[utf8]{inputenc}
\usepackage{color}
\usepackage{xspace}
\usepackage{amsmath}
\usepackage{bbm}
\usepackage{enumitem}
\usepackage{arydshln}
\usepackage{tikz-dependency}

\newcommand{\citet}[1]{\newcite{#1}}  
\newcommand{\citep}[1]{\cite{#1}} 

\newcommand{\Slab}[1]{\label{sec:#1}}
\newcommand{\Sref}[1]{Section~\ref{sec:#1}}
\newcommand{\Tref}[1]{Table~\ref{tab:#1}}
\newcommand{\Fref}[1]{Figure~\ref{fig:#1}}

\newcommand{\KL}{\ensuremath{\mathit{KL}_{\mathit{cpos}^3}}\xspace}
\newcommand{\iKL}{\ensuremath{\mathit{KL}_{\mathit{cpos}^3}^{-4}}\xspace}
\newcommand{\tgt}{\ensuremath{\mathit{tgt}}\xspace}
\newcommand{\src}{\ensuremath{\mathit{src}}\xspace}

\newcommand{\cpos}{\ensuremath{\mathit{cpos}}\xspace}

\DeclareMathOperator{\countt}{count}

\title{Parsing Natural Language Sentences by Semi-supervised Methods}

{\author{Rudolf Rosa \\
Institute of Formal and Applied Linguistics \\
Charles University in Prague, Faculty of Mathematics and Physics\\
Malostranské náměstí 25, Prague, Czech Republic \\
  {\tt rosa@ufal.mff.cuni.cz} \\}

\date{}

\begin{document}
\maketitle
\begin{abstract}
    We present our work on semi-supervised parsing of natural language
    sentences, focusing on multi-source crosslingual transfer of
    delexicalized dependency parsers.
    We first evaluate the influence of treebank annotation styles on parsing
    performance, focusing on adposition attachment style.
    Then, we present \KL, an empirical language similarity measure,
    designed and tuned for source parser weighting in multi-source
    delexicalized parser transfer.
    And finally, we introduce a novel resource
    combination method, based on interpolation of trained parser models.
\end{abstract}

\section{Introduction}

The problem of supervised dependency parsing
of natural language sentences
has been intensively studied for
the past decade, especially since the invention of the graph-based MSTParser
by \citet{mcdonald:2005a}, and the transition-based Malt parser by
\citet{malt}.
The success of these parsing algorithms, together with several CoNLL shared
tasks focused on dependency parsing
\citep{conll:2006,conll:2007,conll:2008,conll:2009},
even lead to a general transition from constituency parsing to dependency parsing
throughout the NLP community.
The current state-of-the-art dependency parsers,
such as the Mate parser of \citet{bohnet:2012},
often achieve around 90\% UAS (Unlabelled Attachment Score) for many languages.

The supervised parsing approaches require labelled training data, i.e.,
manually created dependency treebanks. While these are available for dozens of
languages (see \Sref{data}), only around 1\% of the world's languages are
covered by treebanks. Moreover, treebank annotation is costly, and it is not
expected that most of the remaining languages will be processed any time soon,
or ever. To make matters worse, the existing treebanks necessarily capture texts from
limited domains and limited time periods only, and do not serve us well when
we need to parse texts from a different domain, as shown e.g. by
\citet{gildea2001corpus}.
This naturally motivates research of semi-supervised or unsupervised parsing
methods.

In our work, we focus on semi-supervised approaches to the multilinguality
issue, investigating the
possibilities of using the knowledge contained in treebanks
for one or more source languages (\src)
to analyze sentences of a different target language (\tgt).%
\footnote{While our motivation is the analysis of languages without treebanks, we only
evaluate our methods on languages for which treebanks are available,
and simulate the under-resourced setting by not using the \tgt treebanks for
training.
This is a natural consequence of the fact that without a test treebank, intrinsic
evaluation is impossible, and we are not aware of any reliable scenario for
extrinsic parser evaluation.}
Specifically, we perform transfer of delexicalized dependency parsers -- see
\Sref{parsing}, in which we review the existing approaches.

As noted in \Sref{data}, a plethora of treebank annotation styles
exist, and it is not entirely clear how the annotation style relates to parser
performance. It is well-known that some annotation styles are more easily
learned by dependency parsers than other, but the research in this area is
rather rudimentary even for the monolingual supervised setting, and practically
non-existent in other areas, including cross-lingual parser transfer.
As a prominent example of an annotation difference known to be important for
parser accuracy, but also strongly influencing cross-lingual annotation coherence
(with opposing effects), in \Sref{annotation},
we thoroughly study the appropriateness of two adposition annotation styles,
Prague and Stanford, for delexicalized parser transfer.

Linguistic intuition tells us that for cross-lingual parser transfer, using a
\src treebank of a language very close to the \tgt language should bring the
best results. However, as shown by \citet{mcd:2011}, not only is the
similarity of languages only a weakly established concept,
but the empirical results are often rather counter-intuitive -- for example,
to parse Swedish, the best treebank to use turned out to be a Portuguese one,
performing better than treebanks for Germanic languages (their dataset
included, among other, German, Dutch, Danish, and English).
Therefore, in \Sref{intro:langsim}, we introduce a new empirical language
similarity measure, designed and tuned specifically for the delexicalized
parser transfer approaches, and evaluate its performance in several settings.

Furthermore, in \Sref{modintmain}, we introduce our
own novel method of multisource delexicalized parser transfer, based on
interpolation of trained parser models.
We evaluate the method both in an unweighted as well as a weighted setting,
and compare it to the standard resource combination methods.

Finally, in \Sref{fw}, we present the intention to enrich our approach by
semi-supervised lexicalization in future, which other authors have already shown to have
a great potential of improving the cross-lingual parser transfer performance.

\section{Data}
\Slab{data}

One of the positive side-effects of the CoNLL shared tasks was the assembly of dependency
treebanks for many languages, usually simply referred to as the CoNLL
treebanks, as well as the definition of a file format for
representing parsed sentences -- the CoNLL format.
The datasets were used for evaluation in the shared tasks, but have also become
the de-facto standard for evaluation of later dependency parsers, ensuring
strong comparability of the reported results.

However, the CoNLL treebanks generally use different annotation styles on both
morphological and syntactic level. For example, all treebanks define a POS (part of
speech) tag for each word (or token, more precisely), but the set of POS tags
used by each treebank is different, not only in the level of detail, but
also in the actual tags used to carry the same information -- a noun can be
tagged as n, N, NN, No, S, IZE, etc.
On the syntactic level, not only the sets of labels are different,
but even the unlabelled dependency structures differ, as they correspond to
different linguistic theories; probably the highest variance can be found in
the annotation of coordination structures, as studied by
\citet{biblio:PoMaCoordinationStructures2013}.
While some of the differences may be motivated by inherent properties of the
respective languages, they very often correspond merely to more-or-less
arbitrary design decisions of technical rather than linguistic nature, taken
during the creation of the treebanks.
Importantly, such differences constitute unnecessary obstacles in most multilingual
experiments. For cross-lingual parser transfer, these are absolutely
crucial, leading to low performance of some methods and inapplicability of
other.

The issues with cross-lingually incoherent annotation
first led \citet{interset} to the development of the Interset, a method of
capturing values of most morphological features and for conversions between
various tagsets.
Later, the HamleDT collection of dependency treebanks was created by
\citet{hamledt}, consisting of treebanks harmonized not only on the
morphological level (via Interset), but also on the syntactic level,
loosely following the annotation style of the Prague Dependency Treebank
of \citet{pdt}.

In parallel, \citet{upt} defined the Universal POS tagset (UPOS) as a
counter-weight to Interset, as it only captures 12 most important values of
coarse-grained POS tags, ignoring all other morphological annotation.
It was later used for annotation of the (eventually) 11 treebanks of the Google Universal
Dependency Treebank collection of \citet{mcdonald2013universal}.
For syntactic annotation, the authors defined their own version of the Stanford
Dependencies \citep{sd:2008}, modified to better suit the multilingual
setting, as the original annotation style was implicitly designed for English.
In turn, \citet{sd:2014} reacted by introducing the Universal Stanford
Dependencies as the ``official'' multilingual version of Stanford
Dependencies.
This annotation style was immediately adopted by \citet{hamledt20},
who modified it slightly and used it to ``stanfordize'' the HamleDT collection
by implementing a language-neutral conversion pipeline and applying it to the
harmonized treebanks in HamleDT 2.0.

Recently, all of the harmonization forces have joined together into the
Universal Dependencies project of \citet{udep},%
\footnote{\url{http://universaldependencies.github.io/docs/}}
both defining an annotation style based mainly on UPOS, Interset and Universal Stanford
Dependencies, as well as producing a set of 10 treebanks annotated in this way in the
1.0 version. More treebanks should be available soon, as the 1.1 version is due
on 15th May 2015, and there is a firm plan on continuing to release more
treebanks and to update the annotation style as appropriate in future.
Thus, Universal Dependencies have the ambition of eventually becoming the ultimate
annotation style and dataset for dependency parsing.

\begin{table}
    \centering
    \small
    \begin{tabular}{|ll|r|r|}
        \hline
        & & \multicolumn{2}{c|}{Size (kTokens)}  \\ 
        & Language & Train & Test \\
        \hline
        ar & Arabic & 250 & 28 \\
bg & Bulgarian & 191 & 6 \\
bn & Bengali & 7 & 1 \\
ca & Catalan & 391 & 54 \\
cs & Czech & 1,331 & 174 \\
\hline
da & Danish & 95 & 6 \\
de & German & 649 & 33 \\
el & Greek & 66 & 5 \\
en & English & 447 & 6 \\
es & Spanish & 428 & 51 \\
\hline
et & Estonian & 9 & 1 \\
eu & Basque & 138 & 15 \\
fa & Persian & 183 & 7 \\
fi & Finnish & 54 & 6 \\
grc & Ancient Greek & 304 & 6 \\
\hline
hi & Hindi & 269 & 27 \\
hu & Hungarian & 132 & 8 \\
it & Italian & 72 & 6 \\
ja & Japanese & 152 & 6 \\
la & Latin & 49 & 5 \\
\hline
nl & Dutch & 196 & 6 \\
pt & Portuguese & 207 & 6 \\
ro & Romanian & 34 & 3 \\
ru & Russian & 495 & 4 \\
sk & Slovak & 816 & 86 \\
\hline
sl & Slovenian & 29 & 7 \\
sv & Swedish & 192 & 6 \\
ta & Tamil & 8 & 2 \\
te & Telugu & 6 & 1 \\
tr & Turkish & 66 & 5 \\

        \hline
    \end{tabular}
    \caption{List of HamleDT 2.0 treebanks.}
    \label{tab:langs}
\end{table}

In our work, we carry out all experiments using HamleDT 2.0,
as it is currently still the largest and most diverse harmonized treebank
collection, consisting of 30 treebanks -- see \Tref{langs}.
Specifically, we use the stanfordized version of the collection for most
experiments,%
\footnote{
We chose the Stanford style conversion,
instead of the HamleDT-native Prague style version,
because Stanford Dependencies were developed with
the objective of cross-lingual consistency of dependency structures.
Thus,
we expected them to perform better than other formalisms
in cross-lingual experiments.
Later evaluation of that decision, presented (non-chronologically) in
\Sref{annotation}, showed this assumption to be incorrect, but we have not redone
all our experiments yet with respect to that finding.}
and the gold-standard UPOS tags%
\footnote{More precisely, the tags had been automatically converted from original
gold-standard tags into UPOS tagset with Interset by the authors of HamleDT.}
in all experiments.
%
%
We always use the training sections of the treebanks to train the parsers
or to estimate language similarities, and the test section to evaluate
the methods.

We used 12 of the treebanks
as a development set for hyperparameter tuning
where appropriate to avoid overfitting to the dataset.
The development set consisted of treebanks for Arabic, Bulgarian, Catalan,
Greek, Spanish, Estonian, Persian, Finnish, Hindi, Hungarian, Italian, and
Japanese;%
\footnote{To tune our methods to perform well in many different situations,
we chose the development set to contain both smaller and larger
treebanks,
a pair of very close languages (ca, es),
a very solitary language (ja), multiple members of several language families
(Uralic, Romance),
and both primarily left-branching (bg, el) and right-branching
(ar, ja) languages.}
the remaining 18 treebanks constitute the test set.
For experiments where hyperparameter tuning on the development set was
employed, we report the results of our methods separately for the test set and
for the development set as \tgt treebanks.
However, for each \tgt treebank, be it a test treebank or a development
treebank, all remaining 29 \src treebanks are always used for training in the
evaluation of the methods.

Interestingly, the results of our methods results usually turned out to be generally
similar or better on the test set than on the development set, suggesting that no
overfitting happened. Therefore, we usually discuss both the results on the test set
as well as on the development set when evaluating our experiments.

\section{Delexicalized Parser Transfer}
\Slab{parsing}


In the task of delexicalized dependency parser transfer,
or delex transfer for short,
we train a parser on
a treebank for a resource-rich \src language, using non-lexical
features, most notably
POS tags, but not using word forms or lemmas.
Then, we apply that parser to POS-tagged sentences of a
\tgt language, to obtain a dependency parse tree.
Delexicalized transfer yields worse results than a fully
supervised lexicalized parser, trained on a treebank for the target language.
However, for languages with no treebanks available, 
it may be useful to obtain at least a lower-quality
parse tree for tasks such as information retrieval..

The idea of delexicalized transfer was conceived by
\citet{zeman2008}, who
trained a delexicalized parser on a Danish treebank and evaluated it on
a Swedish one.
They note that while the lexicon of two languages will most
probably differ significantly even if they are very close, they may share many
morphological and syntactic properties. As a prerequisite to applying the
method, they map the treebank POS tagsets to a common set, an approach later
becoming known as conversion to Interset \citep{interset}.
They also normalize the annotation styles of the treebanks to make them
more similar, performing rule-based transformations --
a method that has
developed significantly since then and became known as treebank
harmonization \citep{hamledt}.
We largely build upon all of these approaches in our work.

Usually, multiple \src treebanks are available, and it is non-trivial to
select the best one for a given \tgt language.
Therefore, information from some or all \src treebanks is usually combined
together.
The standard ways are to train a parser on the concatenation of all \src
treebanks (\Sref{tbconcat}),
or to train a separate parser on each \src treebank
and to combine the parse trees produced by the parsers
using a maximum spanning tree algorithm (\Sref{treecomb}).
The tree combination method typically performs better;
it can also be easily extended by weighting the \src parser predictions
by similarity of the \src language to the \tgt language,
which can further improve its results.

\subsection{MSTperl parser}
\Slab{parser}

Throughout this work, we use
the MSTperl parser of \citet{mstperl},
an implementation of
the unlabelled single-best MSTParser of \citet{mcdonald:2005b}, with first-order features and
non-projective parsing.
We train the parser using 3 iterations of MIRA \citep{mira}.%


The MSTParser model uses a set of binary features $F$ that are assigned
weights $w_f$ by training on a treebank.
When parsing a sentence, the parser constructs a complete
weighted directed graph over the tokens of the input
sentence, and assigns each edge $e$ a score $s_e$ which is the sum of weights
of features that are active for that edge:
\begin{equation}
    s_e = \sum_{\forall f \in F} f(e) \cdot w_f \,.
    \label{eq:edgescore}
\end{equation}
The sentence parse tree is the maximum spanning tree (MST)
over that graph, found using the algorithm of \citet{chuliu} and \citet{edmonds}.

Our lexicalized feature set is based on \citep{mcdonald:2005a},
and consists of various conjunctions of the following
features:
\begin{description}
\item [POS tags]
    We use the coarse 12-value UPOS of \citet{upt}.\footnote{%
    These 12 values are:
    NOUN, VERB, ., ADJ, ADP, PRON, CONJ, ADV, PRT, NUM, DET, X.}
    For an edge, we use information about the POS tag of the head, dependent,
    their neighbours, and all of the nodes between them.
\item [Token distance]
    We use signed distance of head and dependent
    ($\mathit{order}_\mathit{head} - \mathit{order}_\mathit{dependent}$),
    bucketed into the following buckets:\\
    $+1;\; +2;\; +3;\; +4;\; \geq\!+5;\; \geq\!+11;$\\
    $-1;\; -2;\; -3;\; -4;\; \leq\!-5;\; \leq\!-11$.
\item [Lexical features]
    We use the word form and the morphological lemma of the head and the dependent.
\end{description}

The delexicalized feature set is based on the lexicalized one, but without the
lexical features.

The usage of only this parser in all experiments somewhat limits the extent of
our findings. Therefore, we intend to employ other parsers in future, e.g. the
Malt parser of \citet{malt}. For most of our approaches, this will be
straightforward, but for the parser model interpolation approach
(\Sref{modintmain}), it may be rather intriguing.

\subsection{Treebank concatenation}
\Slab{tbconcat}

\citet{mcd:2011} applied delexicalized transfer in a
setting with multiple \src treebanks available, finding that
a treebank for a language that is typologically close to the \tgt language
is typically a good choice for the source treebank, but noting that the
problem of selecting the best source treebank is non-trivial;
we will further refer to the best
source treebank as the \emph{oracle treebank}, since it can hardly be
identified without having a \tgt language treebank for evaluation.
As a work-around, the authors proposed a simple resource combination
method -- treebank concatenation -- which consists of the following steps:
\begin{enumerate}[noitemsep]
    \item Concatenate all \src treebanks.
    \item Train a delex parser on the resulting 
        treebank.
    \item Apply the parser to the \tgt text.
\end{enumerate}
Applying this method led to better
results than the average over individual single-source parsers, but worse
than using only the oracle \src parser.
In our work, we take the treebank concatenation method as a
baseline.

\subsection{Parse tree combination}
\Slab{treecomb}

An alternative resource combination approach is the parse tree combination,
used by \citet{sagae2006parser} for monolingual parser
combination.
In this approach, several independent parsers are applied to the same input
sentence, and the parse trees they produce are combined into one
resulting tree.
The combination is performed using the idea of \citet{mcdonald:2005a},
who formulated the problem of finding a parse tree as a problem of finding
the maximum spanning tree of a weighted directed graph of
potential parse tree edges.
In the tree combination method, the weight of each edge is defined as the number of parsers which
include that edge in their output (it can thus also be regarded as a parser
voting approach). To find the MST, we use the
Chu-Liu-Edmonds algorithm \citep{chuliu,edmonds}, which was used by
\citet{mcdonald:2005b} in the non-projective MSTParser.
Other MST algorithms could be
used, such as the Eisner algorithm \citep{eisner}, which is, unlike
Chu-Liu-Edmonds, constrained to producing only projective parse trees,
and was used by \citet{mcdonald:2005a} in the projective MSTParser.

The tree combination method can be easily ported from a monolingual to a
multilingual setting, where the individual parsers are trained over
different languages. 
We take the tree combination method for our base approach to multi-source transfer,
as it yields better results on average than the treebank concatenation method
-- probably because in treebank concatenation,
larger treebanks have more influence on the result,
which may not be well substantiated.

A nice
feature of the tree combination approach is the straightforward
possibility of assigning \emph{weights} to the individual parsers,
as done by \citet{surdeanu:2010} in a
monolingual setting. They let
each parser contribute with a weight based on its performance (attachment
score), thus giving a more powerful vote to parsers that seem to be better on
average.
While this is sensible in a monolingual setting, in multi-source delexicalized
transfer we are more interested in the \emph{language similarity} of the
source and target language, as we would like to give more power to
parsers trained on closer languages (see \Sref{intro:langsim}).

The parse tree combination method proceeds in the following way:
\begin{enumerate}[noitemsep]
    \item Train a delex parser on each \src treebank.
    \item Apply each of the parsers to the \tgt sentence,
        obtaining a set of parse trees.
    \item Construct a weighted directed graph over \tgt sentence tokens,
with each edge assigned a score equal to the number of
parse trees that contain this edge (i.e., each parse tree contributes by 0 or 1 to the
edge score). In the weighted variant, the contribution of
each \src parse tree is multiplied by a weight $w(\tgt, \src)$,
based on language similarity of \tgt and \src.
    \item Find the maximum spanning tree over the graph with the
        Chu-Liu-Edmonds algorithm \citep{chuliu,edmonds}.%
\end{enumerate}


\section{Treebank Annotation Style for Parsing}
\Slab{annotation}

\begin{figure}
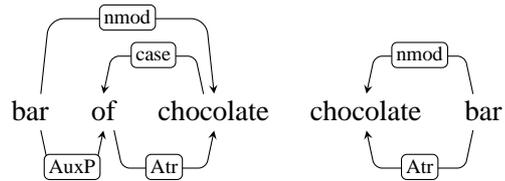

\begin{dependency}
   \begin{deptext}[column sep=1em]
       bar \& of \& chocolate \\
   \end{deptext}
   \depedge{1}{3}{nmod}
   \depedge{3}{2}{case}
   \depedge[edge below]{1}{2}{AuxP}
   \depedge[edge below]{2}{3}{Atr}
\end{dependency}
\begin{dependency}
   \begin{deptext}[column sep=1em]
       chocolate \& bar \\
   \end{deptext}
   \depedge{2}{1}{nmod}
   \depedge[edge below]{2}{1}{Atr}
\end{dependency}
\caption{Stanford style (above) and Prague style (below) analysis of the
phrases ``bar of chocolate'' and ``chocolate bar''. Note that in Stanford
style, these phrases have a more similar structure, both featuring an
\emph{nmod} edge from ``bar'' to ``chocolate''. This shows the principle of
constructions with a similar meaning also having a similar dependency
structure.}
\label{fig:sdprg}
\end{figure}

One of the prominent features of the newest versions of Stanford style
dependencies is their approach to function words. The general rule is that all
function words, such as adpositions or conjunctions, are attached as leaf
nodes. This is a result of a lexicalist view of syntax, which favours
direct dependency relations between lexical nodes directly, not mediated by
function words. This also makes dependency structures more similar
cross-lingually, as it is very common that the same function is expressed by an
adposition in one language, but by other means, such as morphology or word
order, in another language -- or even within the same language, as shown in
\Fref{sdprg}.

The Prague style dependencies, on the other hand, are based
upon a functionalist approach of \citet{sgall1967functional}, and annotate
adpositions as heads of adpositional groups. The lexical nodes are only
directly connected in tectogrammatical (deep-syntax) dependency trees, where function words
are removed and their functions are captured via node attributes.
It is worth noting that in general, there is little
difference between representing information by means of node attributes or
leaf nodes; thus, Stanford trees and Prague tectogrammatical trees are
actually very similar in structure.
However, Prague tectogrammatical trees are rarely directly used in parsing -- they are
typically obtained either by manual annotation, or by automatic conversion
from surface-syntax (analytical) trees.

While Stanford style trees may be more useful for further processing in NLP
applications, it has been argued that Prague style trees are easier to obtain
by using statistical parsers, as, among other differences, adpositions
provide important cues to the parser for adpositional group attachment. This
information becomes harder to access when the adpositions are annotated as
leafs.
The issue of dependency representation learnability has been studied
by several authors, generally reaching similar conclusions
\citep{schwartz2012learnability,sogaard2013empirical,ivanova2013survey}.
The approach suggested by \citet{sd:2014} is to
use a different annotation style for parsing, with Prague style adposition
annotation, among other, and to convert the dependency trees to full Stanford style only
after parsing.

Still, while this seems to be sufficiently proven in the general case, in multi-lingual parsing
scenarios, the higher cross-lingual similarity of Stanford style dependency
trees may be of benefit. From all of the differences between Prague and
Stanford, the adposition attachment seems to be the most important,
as adpositions are usually very frequent and diverse in languages, as well as
very important in parsing.
Therefore, in this section, we evaluate the influence of adposition annotation
style in cross-lingual multi-source delexicalized parser transfer.

In \Sref{fullUSD}, we show that the stanfordized version of HamleDT performs much
worse for parsing than the Prague version.
Consequently, in the following subsections, we use only the Prague version as the basis for
our experiments,
only employing on one of the prominent features of
Stanford Dependencies -- the adposition attachment.
The other annotation differences are currently of less
interest for us, as they concern less frequent phenomena and/or do not seem so
promising for cross-lingual experiments.
Thus, we alternate between Prague adposition attachment as head (denoted ``P''), and
Stanford adposition attachment as leaf node (denoted ``S''),
and thoroughly evaluate the effect of these annotation styles, with a focus on
multi-source delexicalized parser transfer by parse tree combination.

\subsection{Full Universal Stanford Dependencies}
\label{sec:fullUSD}

As a preliminary experiment, we compared the Prague version with its fully stanfordized
version. The results are shown in \Tref{fullUSD}.
It can be seen that the Stanford version performs much worse than the Prague
one -- its results are lower by around 5\% UAS absolute.

\begin{table}
    \centering
    \small
    \begin{tabular}{|l|r|r|r|}
        \hline
        Setup & \multicolumn{1}{c|}{Lex} & \multicolumn{1}{c|}{Delex} & \multicolumn{1}{c|}{Transfer} \\ 
        \hline
        Prague & 80.54 & 74.12 & 56.68 \\ 
        Stanford full & 76.47 & 69.53 & 48.91\\ 
        \hline
        Prague non-punct & 80.23 & 74.00 & 56.08\\ 
        Stanford full non-punct & 76.84 & 70.66& 50.15\\
        \hline
    \end{tabular}
    \caption{Prague versus full Stanford annotation style, UAS averaged over 30 target languages. The
        \emph{Lex}icalized and \emph{Delex}icalized parser is
        monolingual; the \emph{Transfer} parser is a combination of 29 parsers trained on
source treebanks, and then applied to the remaining target language.
Also lists UAS measured only on non-punctuation nodes.}
    \label{tab:fullUSD}
\end{table}

Closer inspection showed that many of the errors are actually due to sentence-final
punctuation attachment. In Stanford style, sentence-final punctuation is to be attached
as a dependent node of the root node of the sentence (typically the main
predicate). However, this is difficult for the first-order parser, as it has
no knowledge of the root node when scoring the potential edges, and thus the
punctuation gets often attached to a different verb.
In Prague style, the sentence-final punctuation is attached to the technical root node,
which is marked by special values of the node features, and thus the
assignment is very easy to make.
While this is an important point to keep in mind when parsing into full
Stanford style,
it is of little relevance to the goal of this paper -- punctuation
attachment is rarely important in NLP applications, and is not very likely to
significantly contribute to cross-lingual dependency structure similarity
either.
For this reason, we also include UAS measured only on non-punctuation nodes.
Still, adposition attachment, which we are mostly interested in, accounts for
only a part of the score difference.

\subsection{Conversion between Prague and Stanford adpositions}
\Slab{prdsdadp}

Further on, we only use the Prague style annotation of the treebanks, with
adpositions annotated either in Prague style (P) or Stanford style (S).
To convert between these adposition annotation styles, we implemented two
simple transformation blocks in the Treex NLP framework \citep{treex}:
\begin{itemize}
    \item The conversion from P to S takes each adposition and attaches it as a
dependent of its left-most non-adpositional child, as well as all of its other
non-adpositional children. Thus, the adposition becomes a leaf node, unless it
has adpositional dependent nodes (typically this signifies a compound adposition).
Coordinating conjunctions are dived through -- if the left-most
non-adpositional child is a coordinating conjunction, the leftmost
non-adpositional conjunct is taken instead (recursively).
    \item In the conversion from S to P, each adposition with a non-adpositional head is
attached as a dependent of its head's head, and its original head is attached
as its dependent.
\end{itemize}
The roundtrip of the conversion (UAS after converting there and back again) is around
98\%.
The transformation blocks,%
\footnote{%
\url{HamleDT::Transform::PrepositionDownwardSimple}
and
\url{HamleDT::Transform::PrepositionUpwardSimple}%
}
as well as the whole Treex framework, are
available on Github.%
\footnote{%
\url{https://github.com/ufal/treex/}\\
\url{https://github.com/ufal/treex/tree/master/lib/Treex/Block/HamleDT/Transform}%
}

Note that there are three places where a conversion from one annotation style to
another may take place -- conversion of the source treebank before training a
parser, conversion of the parser output before the parse tree combination, and
conversion of the parse tree combination output.

\subsection{29-to-1 delexicalized parser transfer}

This section presents and discusses annotation style conversions applied
in semi-supervised parsing of each of the 30 HamleDT 2.0 treebanks as the
\tgt, using delex parsers trained on the remaining 29 \src treebanks,
in an unweighted parse tree combination approach.

As was already mentioned, one of the two adposition
annotation styles (P or S) can be used for parsing, for parse tree combination,
and for converting the output dependency tree.
This yields a number of possible setups, which we will denote as the styles
used for the individual steps,
separated by slashes: parsing/combination/output.

For example, a ``P/S/S'' setup means we use the original P style of the
treebanks for training the parsers, a conversion from P to S of the parse
trees is performed before combining them,
and no conversion of the resulting dependency trees takes place (as they already were in S
style before the combination).
In some of the experiments, we use both P and S parsing -- ``P,S/S/P'' denotes
a setup where both parsers trained on P and S treebanks were applied, the P
style parse trees were then converted to S style, all of these were combined
using the maximum spanning tree algorithm, and the output dependency tree was
then converted to P style.

\begin{table*}
    \centering
    \small
    \begin{tabular}{|l||r|r|r|r|r|r||r|r|r|r|r|r|}
        \hline
        \multicolumn{1}{|c||}{Tgt}
        & \multicolumn{6}{c||}{P style output}
        & \multicolumn{6}{c|}{S style output} \\
        lang & P/P/P & P/S/P & S/S/P & S/P/P & P,S/P/P & P,S/S/P & S/S/S & S/P/S & P/P/S & P/S/S & S,P/S/S & S,P/P/S \\
\hline
ar & 44.61 & 43.07 & 42.29 & 43.32 & \textbf{44.99} & 42.93 & 43.16 & 42.44 & 42.57 & \textbf{44.39} & 44.13 & 43.18 \\
bg & \textbf{73.17} & 71.13 & 70.91 & 71.65 & 72.72 & 71.27 & 72.24 & 72.51 & \textbf{72.68} & 72.55 & 72.65 & 72.58 \\
bn & 59.98 & 59.98 & \textbf{60.47} & \textbf{60.47} & 60.34 & 60.22 & \textbf{60.47} & \textbf{60.47} & 59.98 & 59.98 & 60.22 & 60.34 \\
ca & \textbf{66.45} & 64.78 & 64.49 & 65.32 & 66.38 & 64.73 & 65.61 & 66.08 & 66.10 & \textbf{66.14} & 66.07 & 66.12 \\
cs & 64.06 & 63.21 & 62.94 & 63.50 & \textbf{64.14} & 63.30 & 63.62 & 63.68 & 63.55 & 63.83 & \textbf{63.93} & 63.79 \\
\hline
da & \textbf{63.74} & 61.69 & 62.00 & 62.53 & 63.53 & 61.98 & 62.82 & 62.82 & 62.41 & 62.58 & \textbf{63.09} & 62.41 \\
de & 52.58 & 49.68 & 53.00 & 52.25 & \textbf{55.17} & 52.48 & \textbf{55.95} & 52.52 & 52.18 & 52.40 & 55.32 & 54.92 \\
el & 67.05 & 66.42 & 66.90 & 66.76 & \textbf{67.69} & 67.03 & 67.63 & 67.24 & 66.80 & 67.24 & \textbf{67.78} & 67.63 \\
en & 46.13 & 43.19 & 45.77 & 45.19 & \textbf{48.23} & 45.69 & \textbf{47.65} & 44.31 & 44.69 & 44.41 & 47.09 & 46.71 \\
es & \textbf{69.73} & 67.59 & 67.46 & 68.30 & 69.61 & 67.71 & 68.85 & 69.09 & 69.12 & 69.00 & \textbf{69.17} & 69.09 \\
\hline
et & 71.34 & 70.19 & 71.23 & 71.23 & \textbf{72.07} & 71.76 & 74.06 & 73.12 & 72.59 & 73.85 & \textbf{74.48} & 73.22 \\
eu & 46.12 & \textbf{46.18} & 46.15 & 45.97 & 45.92 & 46.07 & 46.15 & 45.97 & 46.12 & \textbf{46.18} & 46.07 & 45.92 \\
fa & 54.69 & 53.03 & 53.29 & 53.38 & \textbf{54.77} & 53.50 & 56.41 & 55.23 & 55.86 & 56.35 & \textbf{56.69} & 55.84 \\
fi & \textbf{51.48} & 51.34 & 50.47 & 50.40 & 51.17 & 50.99 & 50.60 & 50.54 & \textbf{51.59} & 51.47 & 51.08 & 51.34 \\
grc & 46.24 & 45.50 & 46.29 & \textbf{46.55} & 46.38 & 46.31 & 46.48 & 46.18 & 45.81 & 45.74 & \textbf{46.50} & 45.92 \\
\hline
hi & 30.12 & \textbf{30.66} & 28.96 & 28.41 & 29.64 & 29.60 & 33.23 & 30.81 & 32.10 & \textbf{34.45} & 33.64 & 31.25 \\
hu & 59.68 & \textbf{60.20} & 59.56 & 59.64 & 59.89 & 59.95 & 60.50 & 60.10 & 60.25 & \textbf{61.02} & 60.81 & 60.24 \\
it & 64.52 & 63.60 & 63.97 & 64.60 & \textbf{65.13} & 63.91 & 64.44 & \textbf{64.72} & 64.03 & 64.19 & 64.50 & \textbf{64.72} \\
ja & \textbf{44.23} & 40.94 & 39.36 & 40.27 & 42.64 & 39.87 & 44.02 & 41.55 & 44.02 & \textbf{45.49} & 44.88 & 42.81 \\
la & 41.14 & 40.72 & 41.16 & 40.80 & \textbf{41.28} & 41.22 & 41.34 & 40.93 & 41.28 & 41.05 & \textbf{41.47} & \textbf{41.47} \\
\hline
nl & \textbf{62.47} & 59.09 & 60.63 & 60.72 & 62.04 & 60.39 & \textbf{63.81} & 62.61 & 62.08 & 62.47 & 63.80 & 61.99 \\
pt & 71.35 & 69.78 & 69.97 & 70.97 & \textbf{71.60} & 70.09 & 71.14 & \textbf{71.81} & 70.94 & 71.02 & 71.26 & 71.76 \\
ro & 59.66 & 56.48 & 55.30 & 57.50 & \textbf{59.85} & 55.76 & 58.52 & 58.37 & 59.43 & \textbf{59.51} & 58.67 & 59.39 \\
ru & \textbf{63.82} & 63.36 & 62.20 & 62.78 & 63.65 & 62.84 & 62.43 & 62.87 & \textbf{63.85} & 63.68 & 63.13 & 63.48 \\
sk & 63.66 & 63.02 & 62.96 & 63.22 & \textbf{63.73} & 63.27 & 63.36 & 63.20 & 63.22 & 63.43 & \textbf{63.62} & 63.34 \\
\hline
sl & \textbf{54.40} & 53.35 & 53.16 & 53.24 & 53.68 & 53.15 & 53.80 & 53.18 & \textbf{54.07} & 53.94 & 53.68 & 53.27 \\
sv & 62.08 & 59.00 & 59.97 & 60.87 & \textbf{62.18} & 59.53 & \textbf{62.22} & 61.42 & 60.80 & 60.91 & 61.60 & 61.32 \\
ta & 38.76 & 38.51 & 36.70 & 37.76 & \textbf{39.01} & 38.21 & 37.66 & 37.91 & 38.86 & \textbf{39.06} & 38.91 & \textbf{39.06} \\
te & 66.83 & 66.83 & \textbf{67.00} & 66.83 & 66.16 & 66.50 & \textbf{67.00} & 66.83 & 66.83 & 66.83 & 66.50 & 66.16 \\
tr & 40.39 & 40.35 & 40.77 & 40.66 & 40.82 & \textbf{40.93} & \textbf{41.28} & 40.86 & 40.53 & 40.73 & 41.26 & 40.93 \\
\hline
Avg & 56.68 & 55.43 & 55.51 & 55.84 & \textbf{56.81} & 55.71 & 56.88 & 56.31 & 56.48 & 56.80 & \textbf{57.07} & 56.67 \\

        \hline
    \end{tabular}
    \caption{UAS of various setups of delexicalized parser transfer, always
        using 1 language as target (\emph{Tgt lang}) and the remaining 29 languages as source.
    The best result on each line for both of the output styles is highlighted
in bold. The columns correspond to various combinations of annotation styles
used for parsing/combining/output -- P corresponds to Prague style and S to
Stanford style adposition annotation.}
    \label{tab:main}
\end{table*}

The results are shown in \Tref{main}. For each target language, it shows the
UAS of evaluating various setups on the test section of its treebank.

Clearly, the best results are obtained by parsing both to Stanford and Prague
adposition annotation style, thus obtaining two parse trees for each source
language (58 parse trees for each sentence),
converting the parse trees to the desired output style, and combining them.
This shows that both of the styles have some advantages, and that the parse tree
combination can benefit from these.
Moreover, contrary to supervised parsing, the S style performs better than the
P style, by +0.26\% UAS absolute on average.
This is one of the indications that in this multi-lingual setting, the S style of adposition
attachment is favourable, as it makes the dependency trees more similar.
Overall, using both styles for parsing and Stanford style for combination and
output surpasses the Prague-only baseline by +0.39\% UAS absolute.

A further observation to make is that generally, the P style is good for
parsing, while the S style is good for parse tree combination. Please note the
results of P/P/P and S/P/P, which show a clear dominance (+1.17\%) of using P style for
everything rather than parsing in S and then converting to P.
For the S output style, the difference between S/S/S and P/S/S is only 0.08\%
UAS, and parsing to P and then converting to S and combining actually achieves
the best score for 8 target languages, while using S for everything leads to
best results for only 7 languages.
This can be easily explained, as it has been already shown that the P style is
generally favourable for parsing; the S style then makes the parse trees more
similar and thus easier to combine correctly.

And finally, there is a general tendency of simpler solutions to perform
better -- unless there is a strong benefit of switching styles for a given step, it is
preferable to use a low number of conversions.

\subsection{Smaller source treebank subsets}

For a deeper insight and further confirmation of our conclusions, we also
performed a set of experiments with smaller subsets of the treebank collection.
We selected several treebank groups, based on the ratio of adposition
tokens to all tokens. We also only chose large enough treebanks (more than 100,000
tokens). The subsets are listed in \Tref{seltreebanks}; we also used a
larger ``All9'' set of all the 9 selected treebanks.
Only these were then used for training; the remaining 21 languages were used
for testing as target languages.

\begin{table}
    \centering
    \small
    \begin{tabular}{|l|rl|}
        \hline
        Subset & ADP freq. & Language \\ 
        \hline
             & 15\% & Spanish \\
        High & 19\% & Hindi \\
             & 19\% & Japanese \\
        \hline
             & 9\% & Czech \\
        Med  & 8\% & English \\
             & 9\% & Swedish \\
        \hline
             & 0\% & Basque \\
        Low  & 4\% & Ancient Greek \\
             & 1\% & Hungarian \\
        \hline
             & 15\% & Spanish \\
        Mix  & 9\%  & Swedish \\
             & 1\%  & Hungarian \\
        \hline
    \end{tabular}
    \caption{Subsets of treebanks, selected according to their frequency of
    adposition tokens.}
    \label{tab:seltreebanks}
\end{table}

The summary results are to be found in \Tref{smaller}.
It is easy to see that the conclusions presented in the previous section hold
even for these datasets. Moreover, the differences in UAS are often much
higher, especially for the smaller and highly diverse datasets High, Low and
Mix, where the benefit of the Stanford style making the dependency trees more
similar becomes quite important.
This suggests that the role of Stanford style is stronger with small and
heterogeneous datasets.
For the High dataset, the best result surpasses the Prague-only baseline by as
much as 2.24\% UAS absolute.

\begin{table}
    \centering
    \small
    \begin{tabular}{|l|r|r|r|r|r|}
        \hline
        Setup & \multicolumn{1}{c|}{High}
        & \multicolumn{1}{c|}{Med} & \multicolumn{1}{c|}{Low}
        & \multicolumn{1}{c|}{Mix} & \multicolumn{1}{c|}{All9}  \\ 
        \hline
        P/P/P & 40.53 & 52.00 & 44.53 & 41.03 & 54.98 \\
P/S/P & 39.87 & 50.81 & 41.28 & 38.55 & 52.75 \\
S/S/P & 39.39 & 50.41 & 41.17 & 39.22 & 52.86 \\
S/P/P & 39.68 & 50.86 & 42.15 & 39.36 & 53.79 \\
P,S/P/P & \textbf{41.29} & \textbf{52.57} & \textbf{45.00} & \textbf{41.75} & \textbf{55.37} \\
P,S/S/P & 40.70 & 51.57 & 43.29 & 39.73 & 53.49 \\
\hline
S/S/S & 41.36 & 51.64 & 43.69 & 41.95 & 54.85 \\
S/P/S & 40.43 & 51.49 & 42.97 & 40.66 & 53.88 \\
P/P/S & 40.69 & 51.68 & 44.15 & 40.82 & 54.38 \\
P/S/S & 41.87 & 51.91 & 44.64 & 41.55 & 54.76 \\
S,P/S/S & \textbf{42.77} & \textbf{52.67} & \textbf{46.41} & \textbf{42.66} & \textbf{55.42} \\
S,P/P/S & 41.60 & 52.50 & 44.80 & 41.82 & 54.96 \\

        \hline
    \end{tabular}
    \caption{UAS of delexicalized parser transfer, averaged over 21 languages;
    some or all of the remaining 9 languages are used as source, according to
the given subset name.}
    \label{tab:smaller}
\end{table}

\subsection{Supervised parsers}

\begin{table*}
    \centering
    \small
    \begin{tabular}{|l||r|r|r|r||r|r|r|r||r|r|}
        \hline
        \multicolumn{1}{|c||}{Tgt} & \multicolumn{4}{c||}{Lexicalized supervised}
        & \multicolumn{4}{c||}{Delexicalized supervised}
        & \multicolumn{2}{c|}{Transfer} \\
        \multicolumn{1}{|c||}{lang} & \multicolumn{1}{c|}{P} &
        \multicolumn{1}{c|}{S/P} & \multicolumn{1}{c|}{S} &
        \multicolumn{1}{c||}{P/S} & \multicolumn{1}{c|}{P} &
        \multicolumn{1}{c|}{S/P} & \multicolumn{1}{c|}{S} &
        \multicolumn{1}{c||}{P/S} & \multicolumn{1}{c|}{P,S/P/P} &
        \multicolumn{1}{c|}{S,P/S/S} \\
        \hline
        ar & \textbf{77.47} & 73.92 & 76.32 & 77.17 & \textbf{69.61} & 67.24 & 69.29 & 69.50 & \textbf{44.99} & 44.13 \\
bg & \textbf{87.95} & 85.83 & 87.50 & 87.61 & \textbf{83.87} & 81.11 & 82.76 & 83.32 & \textbf{72.72} & 72.65 \\
bn & 82.27 & \textbf{82.39} & \textbf{82.39} & 82.27 & 77.59 & \textbf{78.82} & \textbf{78.82} & 77.59 & \textbf{60.34} & 60.22 \\
ca & \textbf{86.11} & 81.51 & 84.37 & 85.49 & \textbf{79.71} & 76.41 & 79.03 & 79.33 & \textbf{66.38} & 66.07 \\
cs & \textbf{80.87} & 79.43 & 80.31 & 80.63 & \textbf{70.99} & 70.06 & 70.69 & 70.69 & \textbf{64.14} & 63.93 \\
\hline
da & \textbf{85.66} & 82.55 & 84.42 & 85.12 & \textbf{81.13} & 78.37 & 80.31 & 80.67 & \textbf{63.53} & 63.09 \\
de & \textbf{84.65} & 82.24 & 83.57 & 84.53 & \textbf{77.52} & 75.55 & 76.92 & 77.47 & 55.17 & \textbf{55.32} \\
el & \textbf{80.68} & 79.41 & 80.20 & 80.18 & \textbf{75.40} & 74.10 & 75.15 & 74.73 & 67.69 & \textbf{67.78} \\
en & \textbf{84.71} & 82.85 & 84.37 & 84.05 & \textbf{76.57} & 74.92 & 76.19 & 76.03 & \textbf{48.23} & 47.09 \\
es & \textbf{85.46} & 80.41 & 83.55 & 84.74 & \textbf{79.75} & 75.57 & 78.52 & 79.25 & \textbf{69.61} & 69.17 \\
\hline
et & 85.15 & 85.25 & \textbf{86.30} & 85.46 & 80.96 & 81.38 & \textbf{82.85} & 80.75 & 72.07 & \textbf{74.48} \\
eu & \textbf{75.28} & 75.07 & 75.07 & \textbf{75.28} & 68.34 & \textbf{68.41} & \textbf{68.41} & 68.34 & 45.92 & \textbf{46.07} \\
fa & \textbf{82.27} & 77.94 & 80.21 & 81.70 & 70.44 & 69.17 & \textbf{71.72} & 70.78 & 54.77 & \textbf{56.69} \\
fi & 71.17 & 70.62 & 70.80 & \textbf{71.21} & 63.10 & 62.25 & 62.51 & \textbf{63.13} & \textbf{51.17} & 51.08 \\
grc & \textbf{56.98} & 56.41 & 56.61 & 56.56 & 48.92 & 48.80 & \textbf{49.10} & 48.80 & 46.38 & \textbf{46.50} \\
\hline
hi & 90.40 & 83.92 & 86.43 & \textbf{90.42} & \textbf{80.55} & 78.15 & 80.52 & 80.52 & 29.64 & \textbf{33.64} \\
hu & \textbf{77.60} & 77.03 & 77.07 & 77.40 & \textbf{72.54} & 71.80 & 71.79 & 72.34 & 59.89 & \textbf{60.81} \\
it & \textbf{81.46} & 80.06 & 80.57 & 81.22 & \textbf{77.49} & 76.26 & 76.57 & 76.92 & \textbf{65.13} & 64.50 \\
ja & \textbf{91.17} & 84.54 & 89.65 & 90.79 & 81.72 & 79.83 & 84.03 & \textbf{84.35} & 42.64 & \textbf{44.88} \\
la & 47.55 & 48.69 & \textbf{48.72} & 47.36 & 44.08 & 43.89 & \textbf{44.12} & 43.81 & 41.28 & \textbf{41.47} \\
\hline
nl & \textbf{80.90} & 77.37 & 80.05 & 80.11 & \textbf{74.02} & 70.98 & 73.70 & 73.57 & 62.04 & \textbf{63.80} \\
pt & \textbf{83.50} & 80.62 & 82.21 & 82.97 & \textbf{80.14} & 76.99 & 78.68 & 79.77 & \textbf{71.60} & 71.26 \\
ro & \textbf{89.62} & 86.52 & 88.79 & \textbf{89.62} & 85.19 & 83.41 & \textbf{85.34} & 84.85 & \textbf{59.85} & 58.67 \\
ru & \textbf{83.98} & 82.91 & 83.49 & 83.75 & \textbf{73.08} & 72.24 & 72.70 & 72.90 & \textbf{63.65} & 63.13 \\
sk & \textbf{79.02} & 78.19 & 78.70 & 78.63 & \textbf{71.38} & 70.60 & 70.88 & 70.93 & \textbf{63.73} & 63.62 \\
\hline
sl & \textbf{81.19} & 80.05 & 80.94 & 80.95 & 72.91 & 72.30 & \textbf{72.93} & 72.69 & \textbf{53.68} & \textbf{53.68} \\
sv & \textbf{83.20} & 79.31 & 81.93 & 82.48 & \textbf{78.84} & 75.67 & 77.97 & 78.18 & \textbf{62.18} & 61.60 \\
ta & 72.70 & \textbf{72.75} & 72.60 & 72.30 & \textbf{68.17} & 67.82 & 67.92 & 67.62 & \textbf{39.01} & 38.91 \\
te & \textbf{87.60} & 86.60 & 86.93 & \textbf{87.60} & \textbf{85.59} & 83.75 & 84.09 & 85.59 & 66.16 & \textbf{66.50} \\
tr & \textbf{79.48} & 78.77 & 79.02 & 79.26 & \textbf{73.99} & 73.74 & 73.72 & 73.72 & 40.82 & \textbf{41.26} \\
\hline
Avg & \textbf{80.54} & 78.44 & 79.77 & 80.23 & \textbf{74.12} & 72.65 & 73.91 & 73.94 & 56.81 & \textbf{57.07} \\

        \hline
    \end{tabular}
    \caption{UAS of supervised lexicalized and delexicalized
        monolingual parsers, as well as the two best-performing transfer parser
        setups. For the supervised parsers, the columns corresponds to
    annotation style used for parsing, or parsing/output if a conversion was
performed after parsing. For the delexicalized parser transfer, the columns
correspond to parsing with both P and S style, converting the parse trees to
the same style (P or S), and combining the trees.}
    \label{tab:supervised}
\end{table*}

For completeness, we also include results for supervised monolingual lexicalized
and delexicalized parsers, using the P and S annotation styles of adpositions.
The setup denoted as ``P/S'' corresponds to a parser trained on a P style
target treebank, output of which is converted to S, and then evaluated on the S
conversion of the target treebank test section (analogously for ``S/P'').
For comparison, we also include the two best parser transfer setups (these
results are identical to those in \Tref{main}).

The results are shown in \Tref{supervised}. For the
lexicalized parser, the P style is clearly better, achieving +0.77\% UAS
absolute on average. To obtain S style parse
trees, it is generally better to parse the text using a parser trained on a P
style treebank, and then to convert the output parse trees, which yields a
+0.46\% higher UAS than parsing directly using an S style treebank.
Here, the adpositions clearly provide important information to the parser, and
their annotation as heads benefits the results.

For the delexicalized parser, the P style still performs better (+0.21\% UAS),
although the difference is smaller, and parsing directly using the S style is
comparable to parsing using P style and then converting to S style.
We believe that this is because when word forms and lemmas are removed, the
most important information about the adpositions is missing. If the language
has a general tendency of where it attaches adpositions, the information that
a word is an adposition is still useful, but it has now a limited power
towards adposition attachment disambiguation.

And finally, as has already been discussed, the S style actually performs
better for delexicalized parser transfer than the P style; in the best setups,
the S style achieves +0.26\% UAS on average.

\subsection{Conclusion}

We investigated the usefulness of Stanford adposition attachment
style as an alternative to the Prague style, using a large set of 30 treebanks
for evaluation.
We especially focused on multi-source cross-lingual delexicalized parser transfer, as
one of the targets behind the design of Universal Stanford Dependencies is to
be more cross-lingually consistent than other annotation styles.

We managed to confirm that for supervised parsing, Prague annotation style is
favourable over Stanford style, as has been already stated in literature.
However, in the parser transfer setting, Stanford style adposition attachment
proved to generally perform better than the Prague style, thanks to its
abstraction from the high interlingual variance in adposition usage.
Moreover, even better results are achieved by at once combining outputs of parsers
trained on treebanks of both Prague and Stanford adposition attachment style,
eventually reaching an improvement of +0.39\% UAS absolute over the Prague
style baseline.
Our results are further confirmed by experiments using smaller and more
diverse subsets of training treebanks, where the advantage of Stanford
style often becomes even more pronounced,
reaching an improvement of up to +2.24\% over the Prague style baseline.

In future, we intend to evaluate the effect of other annotation style
differences, such as the coordination structures. We also plan to try to
incorporate more fine-grained morphological information than the UPOS tags,
probably by only including a given feature if it seems to be shared between
the \src and \tgt language, as always including all of them performed very
poorly in preliminary experiments.

\section{Employing Language Similarity}
\Slab{intro:langsim}

The issue of finding a good \src treebank for a given \tgt language can be
approached in two ways. In the single-source approach, we try to find the \src
language which is most similar to the \tgt language, and use the treebank for
that language to train a parser to be applied to the \tgt.
In a multi-source approach, we combine some or all available \src resources,
either in an unweighted way, as was presented in \Sref{parsing},
or weighted by \src-\tgt similarity.
Thus, for both the source selection task and the source weighting task,
there is a need for a language similarity measure, which serves as a proxy for \src
and \tgt treebank similarity, but cannot access the \tgt treebank.
Still, it is reasonable, and usual, to presuppose availability of POS-tagged \tgt
language text (as it constitutes the input to the delex parser),
as well as the information about the identity of the \tgt language.

Several authors \citep{naseem:2012,sogaard:2012,tackstrom:2013}
have employed the World Atlas of Language Structures (WALS) of \citet{wals} 
to estimate the similarity of languages for delex transfer.
They exploit information about the genealogy distance and
shared typological features of the languages, typically word order features.
They note that for a \tgt language which is rather dissimilar to
any of the \src languages, delex transfer achieves better results
if word order is completely or selectively ignored. This is motivated by the
observation that languages, at least when being observed only through POS,
become more similar if we disregard word order.

Apart from using WALS,
Søgaard also takes one other approach to estimating language
similarity.
In \citep{sogaard:2011},
he trains a POS language model on a \tgt POS-tagged
corpus, and uses it to filter the \src treebank, keeping only sentences
that look like target language sentences to the language model. This method
is further improved in
\citep{sogaard:2012}, where the authors move from a selection approach to a
weighting approach: they keep all the \src sentences, but weight each of
them with the score assigned by the language model. This is made possible
by modifying their learning algorithm to use weighted perceptron learning 
\citep{cavallanti:2010}. In this way, they in principle introduce a
weighting scheme for the treebank concatenation.

The central feature of this section is \KL, a language similarity measure based on
similarity of distributions of coarse POS tag trigrams, computed over POS-tagged
corpora for the source and target languages using the Kullback-Leibler
divergence \citep{kl}.
The measure is simple and efficient, does not rely on additional external
resources, and has been designed and tuned specifically to be used in
delexicalized transfer approaches.
We show that \KL performs well in selecting the source treebank in
the single-source delexicalized transfer,
as well as in parser weighting in the multi-source tree combination approach.



\subsection{\KL language similarity measure}

Our method of estimating language similarity for the purposes of delexicalized
transfer is based on comparing distributions of coarse POS trigrams in a
source language treebank ($P_{\src}$) and in a target language POS-tagged
corpus ($P_{\tgt}$).
This is motivated by the fact that POS tags constitute a key feature for
delexicalized parsing.
We use UPOS tags;
we also tried using more fine-grained tags, but this led to
worse results on our development data, probably because
more fine-grained features
tend to be less shared across languages.
We also tried to vary the POS $n$gram length; bigrams and tetragrams both
performed comparably to trigrams on the weighting task, but for the selection
task, trigrams outperformed other $n$grams.

%


The coarse POS trigram distributions are estimated as frequencies computed on the
training parts of the corpora:
\begin{multline}
    f(\cpos_{i-1},\cpos_i,\cpos_{i+1}) = \\
    = \frac{\countt(\cpos_{i-1},\cpos_i,\cpos_{i+1})}{\sum_{ \forall \cpos_{a,b,c} }
    \countt(\cpos_a,\cpos_b,\cpos_c)} \,;
\end{multline}
we use a special value for $\cpos_{i-1}$ or $\cpos_{i+1}$ if $\cpos_i$ appears at
the beginning or end of a sentence, respectively.

We use the Kullback-Leibler divergence%
\footnote{We also tried cosine similarity, with much worse results.}
to compute the similarity
of the distributions as $D_{\text{KL}}(P_{\tgt}||P_{\src})$.\footnote{%
$D_{\text{KL}}(P||Q)$
expresses the amount of information lost when a distribution $Q$ is used to
approximate the true distribution $P$.
Thus, in our setting, we use $D_{\text{KL}}(P_{\tgt}||P_{\src})$,
as we try to minimize the loss of using a parser based on source data
as an approximation of a parser based on the target data.}
The KL divergence of distributions $P$ and $Q$ is defined as
\begin{equation}
    D_{\text{KL}}(P||Q) = \sum_{\forall x} P(x) \cdot \log \frac{P(x)}{Q(x)}\,,
\end{equation}
with the value of the addend defined as $0$ if $P(x) = 0$.
The value of KL divergence is a non-negative number; the more divergent
(dissimilar) the distributions, the higher its value.

In our setting, we compute \KL as
\begin{multline}
    \KL(\tgt, \src) = \\
    = \sum_{\forall \cpos^3 \in \tgt} f_{\tgt}(\cpos^3) \cdot \log
    \frac{f_{\tgt}(\cpos^3)}{f_{\src}(\cpos^3)} \,,
\end{multline}
where $\cpos^3$ is a coarse POS tag trigram. It is sufficient to iterate only over
trigrams present in the target, as the addend is defined to be
zero in other cases. This is in accord with our needs: we do not actually care
about phenomena that the source parser can handle but do not appear in target.

For the KL divergence to be well-defined,
we set the source count of each unseen
trigram to 1.



\subsection{Source selection}

For the single-source parser transfer, we compute \KL distance of the
\tgt corpus to each of the \src treebanks.
We then select the $\src^{*}$ treebank as the closest one:
\begin{equation}
    \src^{*} = \operatorname*{arg\,min}_{\forall \src} \  \KL(\tgt, \src) \,,
\end{equation}
and use it to train the delex parser to be applied to \tgt.

\subsection{Source weighting}
\Slab{weighting}

To convert \KL from a negative measure of language
similarity to a positive \src parser weight, we take
the fourth power of its inverted value, \iKL.
A high value of the exponent strongly promotes the most similar source
language, giving minimal power to the other languages, which is good if
there is a \emph{very} similar source language.
A low value enables combining information from a larger number of
source languages.
We chose a compromise value of $4$ based on performance on the development data.

We then add weighting by \KL into the parse tree combination (\Sref{treecomb})
by multiplying the contribution of each \src parse tree by \iKL(\tgt, \src).

\subsection{Evaluation}
\Slab{langsimeval}

To evaluate the language similarity measure, we use it both for the selection
task and the weighting task on the stanfordized version of the HamleDT 2.0
treebanks.
The exact shape of the measure was tuned on the development set.%
\footnote{We tuned the choice of the similarity measure, POS $n$-gram
length, and the way of turning \KL into \iKL.}

Preliminary trials on the subset of CoNLL 2006 and 2007 data sets
\citep{conll:2006,conll:2007} used by \citet{mcd:2011}
indicated that these are not suitable for our approach, as they are not 
harmonized on the dependency annotation level.\footnote{%
    On the original non-harmonized treebanks, the unweighted tree combination performed
    best (58.06\% UAS), +2.4\% absolute over weighted tree combination and +4.9\%
    over single-source transfer.
    On the harmonized versions of the same treebanks (subset of HamleDT), unweighted tree
    combination was outperformed both by single-source transfer (+1.0\%) and
    weighted tree combination (+1.67\%).
}
There, treebank annotation style similarity seems to become more important than
language similarity; the lack of harmonization makes the data unnecessarily
noisier.

\Tref{kleval} contains the results of our methods
both on the test languages and the
development languages.
For each target language, we used all remaining 29 source languages for
training (in the single-source method, only one of them is selected and applied).
We base our evaluation mainly on average UAS on the test \tgt languages,
and compare the methods by absolute UAS differences.

Our baseline is the treebank concatenation method of
\citet{mcd:2011}, i.e., a single delexicalized parser trained on the
concatenation of the 29 \src treebanks (\Sref{tbconcat}).

As an upper bound,\footnote{%
This is a hard upper-bound for the single-source transfer, but can be
surpassed by the multi-source transfer.}
we report the results of the oracle single-source delexicalized transfer:
for each target language,
the oracle source parser is the one that achieves the highest UAS
on the target treebank test section.\footnote{%
We do not report the matrix of all source/target combination results,
as this amounts to 870 numbers.}
We do not include results of a higher upper bound of
a supervised delexicalized parser (trained on the \tgt treebank), which
has an average UAS of 68.5\%. It was not surpassed by our methods for any
target language, although it was reached for Telugu, and approached within
5\% for Czech and Latin.


\npdecimalsign{.}
\nprounddigits{1}

\begin{table}
    \begin{center}
    \small
    \setlength\tabcolsep{4.6pt}
    \begin{tabular}{|l|r|lr|rlr|rr|}
        \hline
        
        
        \multicolumn{1}{|c|}{Tgt}
        & \multicolumn{1}{c|}{TB}
        & \multicolumn{2}{c|}{Oracle} 
        & \multicolumn{3}{c|}{Single-src}
        & \multicolumn{2}{c|}{Multi-src} \\
        
        \multicolumn{1}{|c|}{lang}
        & \multicolumn{1}{c|}{conc}
        & \multicolumn{2}{c|}{del trans}
        & \multicolumn{1}{c}{KL} & \multicolumn{2}{c|}{}
        & \multicolumn{1}{c|}{$\times 1$} & \multicolumn{1}{c|}{$\times w$} \\
        
        \hline
        \hline

bn & 61.0 & te & \textbf{\np{66.74876847}} & 0.5 & \textbf{te} & \textbf{\np{66.74876847}} & 63.2 & \textbf{\np{66.74876847}} \\
cs & 60.5 & sk & \textbf{\np{65.8261611}} & 0.3 & \textbf{sk} & \textbf{\np{65.8261611}} & 60.4 & \textbf{\np{65.8261611}} \\
da & \textbf{\np{56.16883117}} & en & \textbf{\np{55.43403964}} & 0.5 & sl & 42.1 & \textbf{\np{54.4429255}} & 50.3 \\
\hdashline
de & 12.6 & en & \textbf{\np{56.77270315}} & 0.7 & \textbf{en} & \textbf{\np{56.77270315}} & 27.6 & \textbf{\np{56.77270315}} \\
en & 12.3 & de & \textbf{\np{42.63441935}} & 0.8 & \textbf{de} & \textbf{\np{42.63441935}} & 21.1 & \textbf{\np{42.63441935}} \\
eu & \textbf{\np{41.18594231}} & da & \textbf{\np{42.06104733}} & 0.7 & tr & 29.1 & \textbf{\np{40.78689443}} & 30.6 \\
\hdashline
grc & 43.2 & et & 42.2 & 1.0 & sl & 34.0 & \textbf{\np{44.74701631}} & 42.6 \\
la & 38.1 & grc & 40.3 & 1.2 & cs & 35.0 & \textbf{\np{40.32157027}} & 39.7 \\
nl & 55.0 & da & 57.9 & 0.7 & \textbf{da} & 57.9 & 56.2 & \textbf{\np{58.67502238}} \\
\hdashline
pt & 62.8 & en & 64.2 & 0.2 & es & 62.7 & \textbf{\np{67.15527527}} & 62.7 \\
ro & 44.2 & it & \textbf{\np{66.36363636}} & 1.6 & la & 30.8 & \textbf{\np{51.21212121}} & 50.0 \\
ru & 55.5 & sk & 57.7 & 0.9 & la & 40.4 & \textbf{\np{57.80798149}} & 57.2 \\
\hdashline
sk & 52.2 & cs & \textbf{\np{61.65377682}} & 0.2 & sl & 58.4 & \textbf{\np{59.57318137}} & 58.4 \\
sl & 45.9 & sk & \textbf{\np{53.86541471}} & 0.2 & \textbf{sk} & \textbf{\np{53.86541471}} & 47.1 & \textbf{\np{53.86541471}} \\
sv & 45.4 & de & \textbf{\np{61.61598303}} & 0.6 & da & 49.8 & \textbf{\np{52.29844413}} & 50.8 \\
\hdashline
ta & 27.9 & hi & \textbf{\np{53.54449472}} & 1.1 & tr & 31.1 & 28.0 & \textbf{\np{39.96983409}} \\
te & 67.8 & bn & \textbf{\np{77.38693467}} & 0.4 & \textbf{bn} & \textbf{\np{77.38693467}} & 68.7 & \textbf{\np{77.38693467}} \\
tr & 18.8 & ta & 40.3 & 0.7 & \textbf{ta} & 40.3 & 23.2 & \textbf{\np{41.05916242}} \\
\hline
\textbf{Avg} & 44.5 &  & \textbf{\np{55.9110837}} & 0.7 & \textbf{} & 48.6 & 48.0 & \textbf{\np{52.51526238}} \\
\hdashline
\textbf{SD} & 16.9 &  & 10.8 &  & \textbf{} & 14.4 & 15.0 & 11.8 \\
\hline
\hline
ar & 37.0 & ro & \textbf{\np{43.05430759}} & 1.7 & sk & 41.2 & 35.3 & \textbf{\np{41.3363045}} \\
bg & 64.4 & sk & 66.8 & 0.4 & \textbf{sk} & 66.8 & 66.0 & \textbf{\np{67.37445231}} \\
ca & 56.3 & es & \textbf{\np{72.36819768}} & 0.1 & \textbf{es} & \textbf{\np{72.36819768}} & 61.5 & \textbf{\np{72.36819768}} \\
\hdashline
el & 63.1 & sk & 61.4 & 0.7 & cs & 60.7 & 62.3 & \textbf{\np{63.84263114}} \\
es & 59.9 & ca & \textbf{\np{72.73268742}} & 0.0 & \textbf{ca} & \textbf{\np{72.73268742}} & 64.3 & \textbf{\np{72.73268742}} \\
et & 67.5 & hu & 71.8 & 0.9 & da & 64.9 & 70.5 & \textbf{\np{71.9665272}} \\
\hdashline
fa & 30.9 & ar & \textbf{\np{35.64386017}} & 1.1 & cs & \textbf{\np{34.70271885}} & 32.5 & 33.3 \\
fi & 41.9 & et & 44.2 & 1.1 & \textbf{et} & 44.2 & 41.7 & \textbf{\np{47.13364055}} \\
hi & 24.1 & ta & \textbf{\np{56.26135675}} & 1.1 & fa & 20.8 & 24.6 & \textbf{\np{27.21456693}} \\
\hdashline
hu & 55.1 & et & 52.0 & 0.7 & cs & 46.0 & \textbf{\np{56.54956427}} & 51.2 \\
it & 52.5 & ca & \textbf{\np{59.75274725}} & 0.3 & pt & 54.9 & 59.5 & \textbf{\np{59.61538462}} \\
ja & 29.2 & tr & \textbf{\np{49.23831203}} & 2.2 & ta & \textbf{\np{44.86079496}} & 28.8 & 34.1 \\
\hline
\textbf{Avg} & 48.5 &  & \textbf{\np{57.10675971}} & 0.9 & \textbf{} & 52.0 & 50.3 & \textbf{\np{53.51649087}} \\
\hdashline
\textbf{SD} & 15.2 &  & 12.5 &  & \textbf{} & 16.1 & 16.5 & 16.7 \\
\hline
\hline
\textbf{Avg} & 46.1 &  & \textbf{\np{56.38935411}} & 0.8 & \textbf{} & 50.0 & 48.9 & \textbf{\np{52.91575378}} \\
\hdashline
\textbf{SD} & 16.1 &  & 11.3 &  & \textbf{} & 15.0 & 15.4 & 13.7 \\

        \hline
    \end{tabular}
    \end{center}
    \caption{Evaluation using UAS on test target treebanks (upper part of the
        table) and development target treebanks (lower part).} 
        
        \smallskip\footnotesize
        For each target
        language, all 29 remaining non-target treebanks were used for training
        the parsers. The best score among our transfer methods
        is marked in bold; the baseline and upper bound scores are marked in
        bold if equal to or higher than that.
        
        \smallskip\noindent Legend:\\
        \emph{Tgt lang} = Target treebank language.\\
        \emph{TB conc} = Treebank concatenation.\\
        \emph{Oracle del trans} = Single-source delexicalized
        transfer using the oracle source language.\\
        \emph{Single-src} = Single-source
        delexicalized transfer using source language with lowest \KL
        distance to the target language (language bold if
        identical to oracle).\\
        \emph{Multi-src} = Multi-source delexicalized transfer
        using parse tree combination,
        unweighted ($\times 1$) and \iKL weighted ($\times w$).\\
        \emph{Avg} = Average UAS (on test/development/all).\\
        \emph{SD} = Standard sample deviation of UAS, serving as an
        indication of robustness of the method.
    \label{tab:kleval}
\end{table}

The results show that \KL performs well both in the selection task and in the
weighting task, as both the single-source and the weighted multi-source
transfer methods outperform the unweighted tree combination on average,
as well as the treebank concatenation baseline.
In 8 of 18 cases, \KL is able to correctly identify the oracle source
treebank for the single-source approach. In two of these cases,
weighted tree combination further improves upon the result of the
single-source transfer, i.e., surpasses the oracle;
in the remaining 6 cases, it performs identically to the single-source method.
This proves \KL to be a successful language similarity measure
for delexicalized parser transfer,
and the weighted multi-source transfer to be a better performing approach than the
single-source transfer.

The weighted tree combination is better than its unweighted variant only for
half of the target languages, but it is more stable, as indicated by its lower standard
deviation, and achieves an average UAS higher by 4.5\%.
The unweighted tree combination, as well as treebank concatenation,
perform especially poorly for English, German, Tamil, and
Turkish, which are
rich in determiners,
unlike the rest of the treebanks;\footnote{%
In the treebanks for these four languages, determiners constitute around
5-10\% of all tokens, while most other treebanks contain no
determiners at all; in some cases, this is related to properties of the
treebank annotation or its harmonization rather than properties of the language.}
therefore, determiners are parsed rather
randomly.\footnote{UAS of determiner attachment tends to
be lower than 5\%, which is several times less than for any other POS.}
In the weighted methods, this is not the
case anymore, as for a determiner-rich target language, determiner-rich source
languages are given a high weight.
For target languages for which \KL of the closest source language
was lower or equal to its average
value of 0.7, the oracle treebank was identified in 7 cases out of 12 and a different but
competitive one in 2 cases;
when higher than 0.7,
an appropriate treebank was only chosen in 1
case out of 6.
When \KL failed to identify the oracle, weighted
tree combination was always better or equal to single-source transfer
but mostly worse than unweighted tree combination.
This shows that for distant languages, \KL does not perform as good as for close
languages.


We believe that
taking
multiple characteristics of the languages into account
would improve the results on distant languages.
A good approach might be to use an
empirical measure, such as \KL, combined with supervised information from
other sources, such as WALS.
Alternatively, a backoff approach, i.e. combining \KL with
e.g. $\mathit{KL}_{\mathit{cpos}^2}$, might help to tackle the issue.

Still, for target languages dissimilar to any source language, a better
similarity measure will not help much, as even the oracle results are
usually poor.
More fine-grained resource combination methods are probably needed there,
such as selectively ignoring word order,
or using different sets of weights based on POS of the dependent node.

\subsection{Conclusion}

We presented \KL, an efficient language similarity measure designed for
delexicalized
dependency parser transfer. We evaluated it on a large set of treebanks, and
showed that it performs well in selecting the source
treebank for single-source transfer,
as well as in weighting the source treebanks
in multi-source parse tree combination.

Our method achieves good results when applied to
similar languages, but its performance drops for distant
languages. In future, we intend to explore
combinations of \KL with other language similarity measures,
so that similarity of
distant languages is estimated more reliably.


%


\section{Model Interpolation}
\Slab{modintmain}

In this section, we present a novel method for \src information combination,
based on interpolation of trained MSTperl parser models.
Our approach was motivated by an intuition that the more fine-grained
information provided by the \src edge scores could be of benefit, probably
serving as \src parser confidence.
Moreover, model interpolation is significantly less computationally demanding at inference
than the parse tree combination method,
as instead of running a set of separate \src parsers, only one parser is run.

We are not aware of any prior work on interpolating
dependency parser models; the closest to our approach is the
interpolation of multilingual probabilistic context-free grammars
of \citet{cohen:2011}.


The method proceeds as follows:
\begin{enumerate}[noitemsep]
    \item Train a delex parser model on each \src treebank (\Sref{parser}).
    \item Normalize the parser models (\Sref{normalization}).
    \item Interpolate the parser models (\Sref{modint}, \Sref{wmodint}).
    \item Parse the \tgt text with a delex parser using the interpolated model.
\end{enumerate}

We evaluate the model interpolation method in \Sref{modinteval}, comparing it
both to the treebank concatenation method as well as the parse tree
combination method, in a weighted as well as unweighted setting.

\subsection{Model normalization}
\Slab{normalization}

An important preliminary step to model interpolation is to normalize each of
the trained models,
as the feature weights in models trained over different treebanks
are often not on the same scale
(we do not perform any regularization during the parser training).
We use a simplified version of normalization by standard
deviation. First, we compute the uncorrected sample standard deviation
of the weights of the features in the model
as
\begin{equation}
    s_M = \sqrt{ \frac{1}{|M|} \sum_{\forall f \in M} (w_f - \bar{w})^2 } \,,
\end{equation}
where $\bar{w}$ is the average feature weight,
and $|M|$ is the number of feature weights in model $M$;
only features that were assigned a weight by the training algorithm are taken into account.

We then divide each feature weight by the standard deviation:%
\footnote{We have not found any further gains in performance when subtracting the sample
mean from the weight before the division;
the MSTParser models seem to be typically centered very similarly.}
\begin{equation}
    \forall f \in M: w_f := \frac{w_f}{s_M}\,.
\end{equation}

The choice of normalization by standard deviation is a combination of
its high and stable performance on our development set, and of Occam's razor.
We tried 12 normalization schemes, nearly all of which achieved an
improvement of 2.5\% to 5\% UAS absolute over an interpolation of
unnormalized models on average,
but often with large differences for individual languages.%
\footnote{Another well-performing method was to divide each feature weight by the sum
of absolute values of all feature weights in the model;
or a similar method, applied during inference individually for each sentence,
using only the feature weights that fired for the sentence to compute the divisor.}
%
%




\subsection{Unweighted model interpolation}
\Slab{modint}

The interpolated model is a linear combination of the normalized
models trained over the \src treebanks.
The result is a model that can be used in the same way as a standard MSTperl parser model.

In unweighted model interpolation,
the weight of each feature ($w_f$) is computed as the sum of the weights
of that feature in the normalized \src models ($w_{f,\src}$):
\begin{equation}
    \forall f \in F: w_f = \sum_{\forall \src} w_{f,\src} \,.
    \label{eq:modint}
\end{equation}

\subsection{Weighted model interpolation}
\Slab{wmodint}

In the weighted variant of model interpolation,
we extend (\ref{eq:modint}) with multiplication by a weight $w(\tgt, \src)$,
corresponding to language similarity of \tgt and \src:
\begin{equation}
    \forall f \in F: w_f = \sum_{\forall \src} w_{f,\src} \cdot w(\tgt, \src) \,.
    \label{eq:wmodint}
\end{equation}
In our experiments, we use the $\iKL(\tgt, \src)$ weight,
which we presented in \Sref{intro:langsim}.

\subsection{Evaluation}
\Slab{modinteval}

\npdecimalsign{.}
\nprounddigits{1}

\begin{table}
    \begin{center}
    \small
    \begin{tabular}{|l|rrr|rr|}
        \hline
        
        \multicolumn{1}{|c|}{Target}
        & \multicolumn{3}{c|}{Unweighted}
        & \multicolumn{2}{c|}{Weighted} \\
        
        \multicolumn{1}{|c|}{language}
        & \multicolumn{1}{c|}{Conc}
        & \multicolumn{1}{c|}{Tree}
        & \multicolumn{1}{c|}{Inter}
        & \multicolumn{1}{c|}{Tree}
        & \multicolumn{1}{c|}{Inter} \\
        
        \hline
        \hline

Bengali & 61.0 & 63.2 & \textbf{\np{67.1182266}} & 66.7 & \textbf{\np{66.87192118}} \\
Czech & \textbf{\np{60.48644476}} & 60.4 & 57.5 & \textbf{\np{65.8261611}} & 65.2 \\
Danish & \textbf{\np{56.16883117}} & 54.4 & 48.9 & \textbf{\np{50.3417635}} & 49.5 \\
\hdashline
German & 12.6 & \textbf{\np{27.61527175}} & 18.2 & 56.8 & \textbf{\np{61.6083414}} \\
English & 12.3 & \textbf{\np{21.06735958}} & 16.2 & 42.6 & \textbf{\np{48.59084549}} \\
Basque & \textbf{\np{41.18594231}} & 40.8 & 39.5 & 30.6 & \textbf{\np{34.87118454}} \\
\hdashline
Anc. Greek & 43.2 & \textbf{\np{44.74701631}} & 41.4 & 42.6 & \textbf{\np{44.0073962}} \\
Latin & 38.1 & \textbf{\np{40.32157027}} & 39.7 & \textbf{\np{39.65337231}} & 39.5 \\
Dutch & 55.0 & \textbf{\np{56.23992838}} & 54.2 & 58.7 & \textbf{\np{59.3912265}} \\
\hdashline
Portuguese & 62.8 & \textbf{\np{67.15527527}} & 62.8 & 62.7 & \textbf{\np{63.66115562}} \\
Romanian & 44.2 & \textbf{\np{51.21212121}} & 48.6 & 50.0 & \textbf{\np{50.34090909}} \\
Russian & 55.5 & \textbf{\np{57.80798149}} & 53.3 & \textbf{\np{57.22961249}} & 56.3 \\
\hdashline
Slovak & 52.2 & \textbf{\np{59.57318137}} & 55.7 & 58.4 & \textbf{\np{60.58382276}} \\
Slovenian & 45.9 & \textbf{\np{47.08920188}} & 42.8 & \textbf{\np{53.86541471}} & 49.6 \\
Swedish & 45.4 & \textbf{\np{52.29844413}} & 49.4 & \textbf{\np{50.77793494}} & 50.4 \\
\hdashline
Tamil & 27.9 & \textbf{\np{28.00402212}} & 27.6 & \textbf{\np{39.96983409}} & 37.3 \\
Telugu & 67.8 & 68.7 & \textbf{\np{72.86432161}} & \textbf{\np{77.38693467}} & \textbf{\np{77.38693467}} \\
Turkish & 18.8 & 23.2 & \textbf{\np{25.28251717}} & \textbf{\np{41.05916242}} & 34.8 \\
\hline
\textbf{Average} & 44.5 & \textbf{\np{47.99120762}} & 45.6 & 52.5 & \textbf{\np{52.76828271}} \\
\hdashline
\textbf{Std. dev.} & 16.9 & \textbf{\np{15.0047351}} & 16.0 & \textbf{\np{11.77948133}} & 12.0 \\
\hline
\hline
Arabic & \textbf{\np{36.9622255}} & 35.3 & 30.7 & \textbf{\np{41.3363045}} & 34.6 \\
Bulgarian & 64.4 & \textbf{\np{65.9925851}} & 60.3 & 67.4 & \textbf{\np{68.45298281}} \\
Catalan & 56.3 & \textbf{\np{61.51466566}} & 58.5 & 72.4 & \textbf{\np{72.37008394}} \\
\hdashline
Greek & \textbf{\np{63.13488759}} & 62.3 & 59.6 & 63.8 & \textbf{\np{64.13405495}} \\
Spanish & 59.9 & \textbf{\np{64.28883418}} & 60.4 & 72.7 & \textbf{\np{72.74857052}} \\
Estonian & 67.5 & \textbf{\np{70.50209205}} & 67.4 & \textbf{\np{71.9665272}} & 71.7 \\
\hdashline
Persian & 30.9 & \textbf{\np{32.52166119}} & 29.5 & \textbf{\np{33.313415}} & 28.6 \\
Finnish & \textbf{\np{41.91705069}} & 41.7 & 41.5 & \textbf{\np{47.13364055}} & 44.7 \\
Hindi & 24.1 & 24.6 & \textbf{\np{26.24924288}} & 27.2 & \textbf{\np{32.74152029}} \\
\hdashline
Hungarian & 55.1 & 56.5 & \textbf{\np{57.39379085}} & 51.2 & \textbf{\np{52.95479303}} \\
Italian & 52.5 & \textbf{\np{59.49764521}} & 56.0 & 59.6 & \textbf{\np{60.10596546}} \\
Japanese & \textbf{\np{29.1542637}} & 28.8 & 27.2 & \textbf{\np{34.07459289}} & 33.0 \\
\hline
\textbf{Average} & 48.5 & \textbf{\np{50.28801952}} & 47.9 & \textbf{\np{53.51649087}} & 53.0 \\
\hdashline
\textbf{Std. dev.} & \textbf{\np{15.16088263}} & 16.5 & 15.6 & \textbf{\np{16.71474839}} & 17.4 \\
\hline
\hline
\textbf{Average} & 46.1 & \textbf{\np{48.90993238}} & 46.5 & \textbf{\np{52.91575378}} & 52.9 \\
\hdashline
\textbf{Std. dev.} & 16.1 & \textbf{\np{15.37137314}} & 15.6 & \textbf{\np{13.69530969}} & 14.1 \\

        \hline
    \end{tabular}
    \end{center}
    \caption{UAS on test \tgt treebanks (upper part of
        table) and development \tgt treebanks (lower part).} 
        
        \smallskip\footnotesize
        \emph{Conc} = Treebank concatenation\\
        \emph{Tree} = Parse tree combination\\
        \emph{Inter} = Model interpolation\\
        \emph{Average} = Average UAS (on test/development/all)\\
        \emph{Std. dev.} = Standard sample deviation of UAS, serving as an
        indication of robustness of the method
    \label{tab:modinteval}
\end{table}

\Tref{modinteval} contains the results of our model interpolation methods, as well
as the baseline methods.
For each \tgt language, all remaining 29 \src treebanks were used for parser
training.
We base our evaluation on comparing absolute differences in UAS
on the whole set of 30 languages as targets.%

The performance of the weighted model interpolation is comparable to the weighted
tree combination -- the difference in average UAS of the methods is lower than
0.1\%, with model interpolation achieving a higher UAS than the tree
combination for 16 of the 30 \tgt languages.
This shows that weighted model interpolation is a good alternative to weighted tree
combination.

In the unweighted setting, the situation is quite different, with model
interpolation scoring much lower than tree combination (-2.4\%), and only
slightly higher than treebank concatenation (+0.4\%) on average.
This suggests that, contrary to our original intuition, edge scores assigned
by the \src models are not a good proxy for parser confidence,
not even when appropriately normalized.%
\footnote{The same tendency was observed across all normalization methods
evaluated on the development set.}
Furthermore, the weighted methods generally outperform the unweighted ones
(by +4.0\% for tree combination and by +6.4\% for model interpolation on
average),
which suggests, among other, that the \src-\tgt language similarity is much more
important than the exact values of \src edge scores for resource combination
in delex transfer.

\subsection{Conclusion}

We presented trained parser model interpolation as an alternative method for
multi-source crosslingual delexicalized dependency parser transfer.
Evaluation on a large collection of treebanks showed that
in a setting where the source languages are weighted by their similarity to
the target language, model interpolation performs comparably to the parse tree
combination approach.
Moreover, model interpolation is significantly less computationally demanding
than the tree combination when parsing the target text,
as the interpolation can be efficiently performed beforehand,
thus only requiring to invoke a single parser at runtime,
while in the tree combination approach, each source parser has to be invoked
individually.

In the unweighted setting, model interpolation consistently performed much worse than tree
combination,
which we find rather surprising,
and we therefore plan
to further investigate this in future.
Still, the weighted methods generally outperformed the unweighted ones,
and as the language similarity measure that we used only requires
the source treebanks and a target POS-tagged text,
i.e. exactly the resources that are required even for the unweighted delex transfer
methods, there is little reason not to employ the weighting.
Therefore, the low performance of the unweighted model interpolation is of less
importance than its high performance in the weighted setting.

\section{Cross-lingual Lexicalization}
\Slab{fw}

A very popular method of improving the results of delexicalized parser
transfer is by lexicalizing the
parser in a semi-supervised manner (as manually created parallel
treebanks are extremely rare). 
We did not explore that approach in this work; instead, we focused on improving
the underlying delexicalized parser transfer.
However, we plan to combine it with
semi-supervised lexicalization in future (preliminary experiments
indicate that this leads to further improvements).

\subsection{Employing parallel data}

The typical approach to lexicalization in delex parsing is
by using dictionaries, parallel texts, and/or machine translation techniques
\citep{zhao:mt,mcd:2011,tackstrom:2012,durrett:2012,loganathan:mt}.

One option is to translate the \tgt sentence into the \src language, which
then makes it possible to use a lexicalized \src parser instead of a
delexicalized one. The correspondence of the words in the translation to the
\src words can be established by using word alignment.

If one-to-many or many-to-many alignment is used, the projection of the
syntactic structure through the alignment is non-trivial. Therefore,
word-to-word translation can be used instead.
Another option is to only include the \src information through additional
features, as has been done by \citet{rosa:2012:parser},
which does not require a one-to-one alignment.

If high-quality \src-\tgt parallel texts are available (i.e. created by human
translators, not machine translation systems), they may be used instead to
create an automatic parallel treebank, without the need to use machine
translation (but using a word aligner is still necessary).

A major drawback of all of these methods is the fact that in most cases,
large parallel texts, and thus high-quality machine translation, is only
available for English as the \src language. For some \tgt languages, there may
be other well-resourced \src languages, but generally, and especially for the
focus languages, i.e. under-resourced ones, we are constrained to only using
English \src. As our experiments showed, the English treebank is rarely a good
\src treebank for delex parser transfer; therefore, we expect that the
English-based lexicalization can only serve as a complement to the methods
described in this paper, still using all available \src treebanks.

\subsection{Employing word embeddings}

Recently, especially since the introduction of the word2vec tool by
\citet{mikolov2013efficient}, continuous vector space word representations,
also known as word embeddings, have gained huge popularity, and have proven to
be useful in many tasks of natural language processing.

In our setting, we are especially interested in the approaches that compute
bilingual word vectors, trained to assign similar vectors to words with
similar meaning, regardless of whether these are \src or \tgt language words.
In this way, we could replace the lexical features by embedding features, thus
circumventing the lexicalization problem.

It has to be noted though that our parser, as well as parsers of other
authors, generally support only categorial features; at least in combinations,
but non-combined features are of extremely limited usefulness.
Thus, it is necessary to either convert the vectors from the continuous space
to categorial features, thus losing many of their attractive properties, or to
use a parser that naturally supports combinations of continuous features, which
to the best of our knowledge is not available in present, although we are
aware of ongoing research in this field.

\subsection{Self-training}

Some lexicalization can also be achieved in an unsupervised way by applying
self-training \citep{selftraining}, i.e. parsing a (preferably large)
POS-tagged \tgt corpus by the delex transfer method,
and then using the resulting automatic treebank to train a standard lexicalized
parser in a supervised way.
While it may seem that such a parser has no chance
of outperforming the delex parser, this is not entirely true, as simply the
presence or non-presence of some phenomena in the corpus may help to adjust
some parameters of the parser.
Moreover, for practical reasons, it may be useful to obtain a standard \tgt parser
model to apply for analyzing new \tgt data, rather than always applying the
full multisource transfer machinery.

Our preliminary experiments performed using the training sections of the \tgt
treebanks indicate a small but consistent improvement brought by this approach,
both when training a lexicalized as well as a delexicalized parser on the
automatically parsed data.

The self-training method is quite orthogonal to all the other approaches, and
can thus presumably be applied on top of any future parsing system.

%

\section{Conclusion}

We presented our work on multi-source crosslingual transfer of
delexicalized dependency parsers.

We evaluated the influence of treebank annotation styles on parsing
performance, focusing on adposition attachment style, and found that the
Stanford annotation, while being infavourable for supervised parsers, performs
promisingly in the multilingual delexicalized parser transfer setting.

We then presented \KL, an empirical language similarity measure
designed for source parser weighting in multi-source
delexicalized parser transfer. We demonstrated its generally good performance,
although improvements still have to be made for cases where the target
language is too dissimilar to any available source language.

And finally, we introduced a novel resource
combination method, based on interpolation of trained MSTParser models.
Although we found its performance to be below our expectations, when combined
with \iKL weighting, its results match these of the weighted parse tree combination
method. As model interpolation is less computationally demanding than parse
tree combination, we find it to be a good alternative multisource
delexicalized parser transfer method.

Throughout our work, we also identified numerous promising paths for further
research, the most important being semi-supervised lexicalization of the methods.

\section*{Acknowledgments}
This research was supported by the grants
GAUK 1572314 
and SVV~260~224.
This work has been using language resources developed,
stored and distributed by the LINDAT/\discretionary{}{}{}CLARIN project of the 
Ministry of Education, Youth and Sports of the Czech Republic (project LM2010013).

\bibliographystyle{acl}
\bibliography{references}

\begin{thebibliography}{}

\bibitem[\protect\citename{B{\"o}hmov{\'a} \bgroup et al.\egroup }2003]{pdt}
Alena B{\"o}hmov{\'a}, Jan Haji{\v{c}}, Eva Haji{\v{c}}ov{\'a}, and Barbora
  Hladk{\'a}.
\newblock 2003.
\newblock The prague dependency treebank.
\newblock In {\em Treebanks}, pages 103--127. Springer.

\bibitem[\protect\citename{Bohnet and Nivre}2012]{bohnet:2012}
Bernd Bohnet and Joakim Nivre.
\newblock 2012.
\newblock A transition-based system for joint part-of-speech tagging and
  labeled non-projective dependency parsing.
\newblock In {\em Proceedings of the 2012 Joint Conference on Empirical Methods
  in Natural Language Processing and Computational Natural Language Learning},
  EMNLP-CoNLL '12, pages 1455--1465, Stroudsburg, PA, USA. Association for
  Computational Linguistics.

\bibitem[\protect\citename{Buchholz and Marsi}2006]{conll:2006}
Sabine Buchholz and Erwin Marsi.
\newblock 2006.
\newblock {CoNLL-X} shared task on multilingual dependency parsing.
\newblock In {\em Proceedings of the Tenth Conference on Computational Natural
  Language Learning}, pages 149--164. Association for Computational
  Linguistics.

\bibitem[\protect\citename{Cavallanti \bgroup et al.\egroup
  }2010]{cavallanti:2010}
Giovanni Cavallanti, Nicolo Cesa-Bianchi, and Claudio Gentile.
\newblock 2010.
\newblock Linear algorithms for online multitask classification.
\newblock {\em The Journal of Machine Learning Research}, 11:2901--2934.

\bibitem[\protect\citename{Chu and Liu}1965]{chuliu}
Yoeng-Jin Chu and Tseng-Hong Liu.
\newblock 1965.
\newblock On shortest arborescence of a directed graph.
\newblock {\em Scientia Sinica}, 14(10):1396.

\bibitem[\protect\citename{Cohen \bgroup et al.\egroup }2011]{cohen:2011}
Shay~B. Cohen, Dipanjan Das, and Noah~A. Smith.
\newblock 2011.
\newblock Unsupervised structure prediction with non-parallel multilingual
  guidance.
\newblock In {\em Proceedings of the Conference on Empirical Methods in Natural
  Language Processing}, EMNLP '11, pages 50--61, Stroudsburg, PA, USA.
  Association for Computational Linguistics.

\bibitem[\protect\citename{Crammer and Singer}2003]{mira}
Koby Crammer and Yoram Singer.
\newblock 2003.
\newblock Ultraconservative online algorithms for multiclass problems.
\newblock {\em The Journal of Machine Learning Research}, 3:951--991.

\bibitem[\protect\citename{De~Marneffe and Manning}2008]{sd:2008}
Marie-Catherine De~Marneffe and Christopher~D Manning.
\newblock 2008.
\newblock The stanford typed dependencies representation.
\newblock In {\em Coling 2008: Proceedings of the workshop on Cross-Framework
  and Cross-Domain Parser Evaluation}, pages 1--8. Association for
  Computational Linguistics.

\bibitem[\protect\citename{de Marneffe \bgroup et al.\egroup }2014]{sd:2014}
Marie-Catherine de~Marneffe, Natalia Silveira, Timothy Dozat, Katri Haverinen,
  Filip Ginter, Joakim Nivre, and Christopher~D. Manning.
\newblock 2014.
\newblock Universal {Stanford} dependencies: A cross-linguistic typology.
\newblock In {\em Proc. of LREC'14}, Reykjavík, Iceland. European Language
  Resources Association (ELRA).

\bibitem[\protect\citename{Dryer and Haspelmath}2013]{wals}
Matthew~S. Dryer and Martin Haspelmath, editors.
\newblock 2013.
\newblock {\em WALS Online}.
\newblock Max Planck Institute for Evolutionary Anthropology, Leipzig.

\bibitem[\protect\citename{Durrett \bgroup et al.\egroup }2012]{durrett:2012}
Greg Durrett, Adam Pauls, and Dan Klein.
\newblock 2012.
\newblock Syntactic transfer using a bilingual lexicon.
\newblock In {\em Proceedings of the 2012 Joint Conference on Empirical Methods
  in Natural Language Processing and Computational Natural Language Learning},
  pages 1--11. Association for Computational Linguistics.

\bibitem[\protect\citename{Edmonds}1967]{edmonds}
Jack Edmonds.
\newblock 1967.
\newblock Optimum branchings.
\newblock {\em Journal of Research of the National Bureau of Standards B},
  71(4):233--240.

\bibitem[\protect\citename{Eisner}1996]{eisner}
Jason~M. Eisner.
\newblock 1996.
\newblock Three new probabilistic models for dependency parsing: An
  exploration.
\newblock In {\em Proceedings of the 16th Conference on Computational
  Linguistics - Volume 1}, COLING '96, pages 340--345, Stroudsburg, PA, USA.
  Association for Computational Linguistics.

\bibitem[\protect\citename{Gildea}2001]{gildea2001corpus}
Daniel Gildea.
\newblock 2001.
\newblock Corpus variation and parser performance.
\newblock In {\em Proceedings of the 2001 Conference on Empirical Methods in
  Natural Language Processing}, pages 167--202.

\bibitem[\protect\citename{Haji{\v{c}} \bgroup et al.\egroup }2009]{conll:2009}
Jan Haji{\v{c}}, Massimiliano Ciaramita, Richard Johansson, Daisuke Kawahara,
  Maria~Ant{\`o}nia Mart{\'\i}, Llu{\'\i}s M{\`a}rquez, Adam Meyers, Joakim
  Nivre, Sebastian Pad{\'o}, Jan {\v{S}}t{\v{e}}p{\'a}nek, et~al.
\newblock 2009.
\newblock The conll-2009 shared task: Syntactic and semantic dependencies in
  multiple languages.
\newblock In {\em Proceedings of the Thirteenth Conference on Computational
  Natural Language Learning: Shared Task}, pages 1--18. Association for
  Computational Linguistics.

\bibitem[\protect\citename{Ivanova \bgroup et al.\egroup
  }2013]{ivanova2013survey}
Angelina Ivanova, Stephan Oepen, and Lilja {\O}vrelid.
\newblock 2013.
\newblock Survey on parsing three dependency representations for english.
\newblock In {\em ACL (Student Research Workshop)}, pages 31--37.

\bibitem[\protect\citename{Kullback and Leibler}1951]{kl}
Solomon Kullback and Richard~A Leibler.
\newblock 1951.
\newblock On information and sufficiency.
\newblock {\em The annals of mathematical statistics}, pages 79--86.

\bibitem[\protect\citename{McClosky \bgroup et al.\egroup }2006]{selftraining}
David McClosky, Eugene Charniak, and Mark Johnson.
\newblock 2006.
\newblock Effective self-training for parsing.
\newblock In {\em Proceedings of the Main Conference on Human Language
  Technology Conference of the North American Chapter of the Association of
  Computational Linguistics}, HLT-NAACL '06, pages 152--159, Stroudsburg, PA,
  USA. Association for Computational Linguistics.

\bibitem[\protect\citename{McDonald \bgroup et al.\egroup
  }2005a]{mcdonald:2005a}
Ryan McDonald, Koby Crammer, and Fernando Pereira.
\newblock 2005a.
\newblock Online large-margin training of dependency parsers.
\newblock In {\em Proceedings of the 43rd annual meeting on association for
  computational linguistics}, pages 91--98. Association for Computational
  Linguistics.

\bibitem[\protect\citename{McDonald \bgroup et al.\egroup
  }2005b]{mcdonald:2005b}
Ryan McDonald, Fernando Pereira, Kiril Ribarov, and Jan Haji{\v{c}}.
\newblock 2005b.
\newblock Non-projective dependency parsing using spanning tree algorithms.
\newblock In {\em Proceedings of the conference on Human Language Technology
  and Empirical Methods in Natural Language Processing}, pages 523--530.
  Association for Computational Linguistics.

\bibitem[\protect\citename{McDonald \bgroup et al.\egroup }2011]{mcd:2011}
Ryan McDonald, Slav Petrov, and Keith Hall.
\newblock 2011.
\newblock Multi-source transfer of delexicalized dependency parsers.
\newblock In {\em Proceedings of the Conference on Empirical Methods in Natural
  Language Processing}, EMNLP '11, pages 62--72, Stroudsburg, PA, USA.
  Association for Computational Linguistics.

\bibitem[\protect\citename{McDonald \bgroup et al.\egroup
  }2013]{mcdonald2013universal}
Ryan~T McDonald, Joakim Nivre, Yvonne Quirmbach-Brundage, Yoav Goldberg,
  Dipanjan Das, Kuzman Ganchev, Keith~B Hall, Slav Petrov, Hao Zhang, Oscar
  T{\"a}ckstr{\"o}m, et~al.
\newblock 2013.
\newblock Universal dependency annotation for multilingual parsing.
\newblock In {\em ACL (2)}, pages 92--97.

\bibitem[\protect\citename{Mikolov \bgroup et al.\egroup
  }2013]{mikolov2013efficient}
Tomas Mikolov, Kai Chen, Greg Corrado, and Jeffrey Dean.
\newblock 2013.
\newblock Efficient estimation of word representations in vector space.
\newblock {\em Proceedings of Workshop at ICLR}.

\bibitem[\protect\citename{Naseem \bgroup et al.\egroup }2012]{naseem:2012}
Tahira Naseem, Regina Barzilay, and Amir Globerson.
\newblock 2012.
\newblock Selective sharing for multilingual dependency parsing.
\newblock In {\em Proceedings of the 50th Annual Meeting of the Association for
  Computational Linguistics: Long Papers - Volume 1}, ACL '12, pages 629--637,
  Stroudsburg, PA, USA. Association for Computational Linguistics.

\bibitem[\protect\citename{Nilsson \bgroup et al.\egroup }2007]{conll:2007}
Jens Nilsson, Sebastian Riedel, and Deniz Yuret.
\newblock 2007.
\newblock The {CoNLL} 2007 shared task on dependency parsing.
\newblock In {\em Proceedings of the CoNLL shared task session of EMNLP-CoNLL},
  pages 915--932. sn.

\bibitem[\protect\citename{Nivre \bgroup et al.\egroup }2006]{malt}
Joakim Nivre, Johan Hall, and Jens Nilsson.
\newblock 2006.
\newblock Maltparser: A data-driven parser-generator for dependency parsing.
\newblock In {\em Proceedings of LREC}.

\bibitem[\protect\citename{Nivre \bgroup et al.\egroup }2015]{udep}
Joakim Nivre, Cristina Bosco, Jinho Choi, Marie-Catherine de~Marneffe, Timothy
  Dozat, Rich{\'a}rd Farkas, Jennifer Foster, Filip Ginter, Yoav Goldberg, Jan
  Haji{\v c}, Jenna Kanerva, Veronika Laippala, Alessandro Lenci, Teresa Lynn,
  Christopher Manning, Ryan {McDonald}, Anna Missil{\"a}, Simonetta Montemagni,
  Slav Petrov, Sampo Pyysalo, Natalia Silveira, Maria Simi, Aaron Smith, Reut
  Tsarfaty, Veronika Vincze, and Daniel Zeman.
\newblock 2015.
\newblock Universal dependencies 1.0.

\bibitem[\protect\citename{Petrov \bgroup et al.\egroup }2012]{upt}
Slav Petrov, Dipanjan Das, and Ryan McDonald.
\newblock 2012.
\newblock A universal part-of-speech tagset.
\newblock In {\em Proc. of LREC-2012}, pages 2089--2096, Istanbul, Turkey.
  European Language Resources Association (ELRA).

\bibitem[\protect\citename{Popel \bgroup et al.\egroup
  }2013]{biblio:PoMaCoordinationStructures2013}
Martin Popel, David Mare{\v{c}}ek, Jan {\v{S}}t{\v{e}}p{\'{a}}nek, Daniel
  Zeman, and Zden{\v{e}}k {\v{Z}}abokrtsk{\'{y}}.
\newblock 2013.
\newblock Coordination structures in dependency treebanks.
\newblock In {\em Proceedings of the 51st Annual Meeting of the Association for
  Computational Linguistics}, pages 517--527, Sofija, Bulgaria. B{\u{a}}lgarska
  akademija na naukite, Association for Computational Linguistics.

\bibitem[\protect\citename{Ramasamy \bgroup et al.\egroup }2014]{loganathan:mt}
Loganathan Ramasamy, David Mare{\v{c}}ek, and Zden{\v{e}}k
  {\v{Z}}abokrtsk{\'{y}}.
\newblock 2014.
\newblock Multilingual dependency parsing: Using machine translated texts
  instead of parallel corpora.
\newblock {\em The Prague Bulletin of Mathematical Linguistics}, 102:93--104.

\bibitem[\protect\citename{Rosa \bgroup et al.\egroup }2012]{rosa:2012:parser}
Rudolf Rosa, Ond{\v{r}}ej Du{\v{s}}ek, David Mare{\v{c}}ek, and Martin Popel.
\newblock 2012.
\newblock Using parallel features in parsing of machine-translated sentences
  for correction of grammatical errors.
\newblock In {\em Proceedings of Sixth Workshop on Syntax, Semantics and
  Structure in Statistical Translation ({SSST}-6), {ACL}}, pages 39--48, Jeju,
  Korea. {ACL}.

\bibitem[\protect\citename{Rosa \bgroup et al.\egroup }2014]{hamledt20}
Rudolf Rosa, Jan Ma{\v{s}}ek, David Mare{\v{c}}ek, Martin Popel, Daniel Zeman,
  and Zden{\v{e}}k {\v{Z}}abokrtsk{\'{y}}.
\newblock 2014.
\newblock Hamle{DT} 2.0: Thirty dependency treebanks stanfordized.
\newblock In Nicoletta Calzolari, Khalid Choukri, Thierry Declerck, Hrafn
  Loftsson, Bente Maegaard, and Joseph Mariani, editors, {\em Proceedings of
  the 9th International Conference on Language Resources and Evaluation ({LREC}
  2014)}, pages 2334--2341, Reykjav{\'{i}}k, Iceland. European Language
  Resources Association.

\bibitem[\protect\citename{Rosa}2014]{mstperl}
Rudolf Rosa.
\newblock 2014.
\newblock {MST}perl parser.

\bibitem[\protect\citename{Sagae and Lavie}2006]{sagae2006parser}
Kenji Sagae and Alon Lavie.
\newblock 2006.
\newblock Parser combination by reparsing.
\newblock In {\em Proceedings of the Human Language Technology Conference of
  the NAACL, Companion Volume: Short Papers}, pages 129--132. Association for
  Computational Linguistics.

\bibitem[\protect\citename{Schwartz \bgroup et al.\egroup
  }2012]{schwartz2012learnability}
Roy Schwartz, Omri Abend, and Ari Rappoport.
\newblock 2012.
\newblock Learnability-based syntactic annotation design.
\newblock In {\em Proceedings of COLING 2012: Technical Papers}.

\bibitem[\protect\citename{Sgall}1967]{sgall1967functional}
Petr Sgall.
\newblock 1967.
\newblock Functional sentence perspective in a generative description.
\newblock {\em Prague studies in mathematical linguistics}, 2(203-225).

\bibitem[\protect\citename{S{\o}gaard and Wulff}2012]{sogaard:2012}
Anders S{\o}gaard and Julie Wulff.
\newblock 2012.
\newblock An empirical etudy of non-lexical extensions to delexicalized
  transfer.
\newblock In {\em COLING (Posters)}, pages 1181--1190.

\bibitem[\protect\citename{S{\o}gaard}2011]{sogaard:2011}
Anders S{\o}gaard.
\newblock 2011.
\newblock Data point selection for cross-language adaptation of dependency
  parsers.
\newblock In {\em Proceedings of the 49th Annual Meeting of the Association for
  Computational Linguistics: Human Language Technologies: short papers-Volume
  2}, pages 682--686. Association for Computational Linguistics.

\bibitem[\protect\citename{S{\o}gaard}2013]{sogaard2013empirical}
Anders S{\o}gaard.
\newblock 2013.
\newblock An empirical study of differences between conversion schemes and
  annotation guidelines.
\newblock In {\em Proceedings of the Second International Conference on
  Dependency Linguistics (DepLing 2013), Prague, Czech Republic: Charles
  University in Prague, Matfyzpress}, pages 298--307.

\bibitem[\protect\citename{Surdeanu and Manning}2010]{surdeanu:2010}
Mihai Surdeanu and Christopher~D. Manning.
\newblock 2010.
\newblock Ensemble models for dependency parsing: Cheap and good?
\newblock In {\em Human Language Technologies: The 2010 Annual Conference of
  the North American Chapter of the Association for Computational Linguistics},
  HLT '10, pages 649--652, Stroudsburg, PA, USA. Association for Computational
  Linguistics.

\bibitem[\protect\citename{Surdeanu \bgroup et al.\egroup }2008]{conll:2008}
Mihai Surdeanu, Richard Johansson, Adam Meyers, Llu{\'\i}s M{\`a}rquez, and
  Joakim Nivre.
\newblock 2008.
\newblock The conll-2008 shared task on joint parsing of syntactic and semantic
  dependencies.
\newblock In {\em Proceedings of the Twelfth Conference on Computational
  Natural Language Learning}, pages 159--177. Association for Computational
  Linguistics.

\bibitem[\protect\citename{T{\"a}ckstr{\"o}m \bgroup et al.\egroup
  }2012]{tackstrom:2012}
Oscar T{\"a}ckstr{\"o}m, Ryan McDonald, and Jakob Uszkoreit.
\newblock 2012.
\newblock Cross-lingual word clusters for direct transfer of linguistic
  structure.
\newblock In {\em Proceedings of the 2012 Conference of the North American
  Chapter of the Association for Computational Linguistics: Human Language
  Technologies}, pages 477--487. Association for Computational Linguistics.

\bibitem[\protect\citename{T{\"a}ckstr{\"o}m \bgroup et al.\egroup
  }2013]{tackstrom:2013}
Oscar T{\"a}ckstr{\"o}m, Ryan McDonald, and Joakim Nivre.
\newblock 2013.
\newblock Target language adaptation of discriminative transfer parsers.

\bibitem[\protect\citename{\v{Z}abokrtsk{\'y}}2011]{treex}
Zden\v{e}k \v{Z}abokrtsk{\'y}.
\newblock 2011.
\newblock Treex-an open-source framework for natural language processing.
\newblock In {\em ITAT}, pages 7--14.

\bibitem[\protect\citename{Zeman and Resnik}2008]{zeman2008}
Daniel Zeman and Philip Resnik.
\newblock 2008.
\newblock Cross-language parser adaptation between related languages.
\newblock In {\em {IJCNLP} 2008 Workshop on {NLP} for Less Privileged
  Languages}, pages 35--42, Hyderabad, India. Asian Federation of Natural
  Language Processing, International Institute of Information Technology.

\bibitem[\protect\citename{Zeman \bgroup et al.\egroup }2012]{hamledt}
Daniel Zeman, David Mareček, Martin Popel, Loganathan Ramasamy, Jan
  Štěpánek, Zdeněk Žabokrtský, and Jan Hajič.
\newblock 2012.
\newblock Hamledt: To parse or not to parse?
\newblock In Nicoletta Calzolari~(Conference Chair), Khalid Choukri, Thierry
  Declerck, Mehmet~Uğur Doğan, Bente Maegaard, Joseph Mariani, Jan Odijk, and
  Stelios Piperidis, editors, {\em Proceedings of the Eight International
  Conference on Language Resources and Evaluation (LREC'12)}, Istanbul, Turkey,
  May. European Language Resources Association (ELRA).

\bibitem[\protect\citename{Zeman}2008]{interset}
Daniel Zeman.
\newblock 2008.
\newblock Reusable tagset conversion using tagset drivers.
\newblock In {\em Proceedings of the 6th International Conference on Language
  Resources and Evaluation ({LREC} 2008)}, pages 213--218, Marrakech, Morocco.
  European Language Resources Association.

\bibitem[\protect\citename{Zhao \bgroup et al.\egroup }2009]{zhao:mt}
Hai Zhao, Yan Song, Chunyu Kit, and Guodong Zhou.
\newblock 2009.
\newblock Cross language dependency parsing using a bilingual lexicon.
\newblock In {\em Proceedings of the Joint Conference of the 47th Annual
  Meeting of the ACL and the 4th International Joint Conference on Natural
  Language Processing of the AFNLP: Volume 1 - Volume 1}, ACL '09, pages
  55--63, Stroudsburg, PA, USA. Association for Computational Linguistics.

\end{thebibliography}

\end{document}